\documentclass{article}
\usepackage[preprint]{tmlr}

\usepackage{amsmath}
\usepackage{amssymb}
\usepackage{booktabs}
\usepackage{multirow}
\usepackage{longtable}
\usepackage{float}
\usepackage{graphicx}
\usepackage{xcolor}
\usepackage{algorithm}

\usepackage{algorithmic}
\usepackage{enumitem}
\usepackage{tcolorbox}
\usepackage{hyperref}
\usepackage{url}
\definecolor{tmlrlink}{RGB}{0,60,120}
\hypersetup{colorlinks=true, linkcolor=tmlrlink, citecolor=tmlrlink, urlcolor=tmlrlink}

\newcommand{\IR}{\mathrm{FS}}
\newcommand{\RNaclTotal}{69{,}717}
\newcommand{\RNbh}{65}
\newcommand{\RNbhMal}{34}
\newcommand{\RNbhPfinalJoint}{30}
\newcommand{\RNchunkAMal}{1.067}
\newcommand{\RNchunkAMalPCount}{10}
\newcommand{\RNchunkAMalPSig}{2}
\newcommand{\RNchunkARw}{0.593}
\newcommand{\RNchunkARwPCount}{10}
\newcommand{\RNchunkARwPSig}{0}
\newcommand{\RNchunkATypx}{0.463}
\newcommand{\RNchunkATypxPCount}{5}
\newcommand{\RNchunkATypxPSig}{0}
\newcommand{\RNchunkBMal}{1.239}
\newcommand{\RNchunkBRw}{0.689}
\newcommand{\RNchunkBTypx}{0.538}
\newcommand{\RNciShrink}{19}
\newcommand{\RNciShrinkMO}{54}
\newcommand{\RNclusterRob}{26}
\newcommand{\RNewmaCIhi}{0.115}
\newcommand{\RNewmaCIlo}{0.070}
\newcommand{\RNewmaMOhi}{0.132}
\newcommand{\RNewmaMOlo}{0.053}
\newcommand{\RNewmaN}{11{,}395}
\newcommand{\RNewmaNaivHi}{0.111}
\newcommand{\RNewmaNaivLo}{0.074}
\newcommand{\RNewmaNeff}{7{,}503}
\newcommand{\RNewmaNeffMO}{2{,}390}
\newcommand{\RNewmaR}{0.093}
\newcommand{\RNewmaRho}{0.653}
\newcommand{\RNexplConv}{152}
\newcommand{\RNhiBH}{20}
\newcommand{\RNhiIR}{65.0}
\newcommand{\RNhiRob}{7}
\newcommand{\RNholmAConfirmed}{2}
\newcommand{\RNholmBConfirmed}{2}
\newcommand{\RNirBHJoint}{46.2}
\newcommand{\RNirMeld}{9.1}
\newcommand{\RNirRawJoint}{47.8}
\newcommand{\RNirRawMO}{47.8}
\newcommand{\RNirSwbd}{15.8}
\newcommand{\RNlabMal}{17}
\newcommand{\RNlabRw}{9}
\newcommand{\RNlabTypx}{9}
\newcommand{\RNlouoHi}{67.5}
\newcommand{\RNlouoLo}{43.3}
\newcommand{\RNlowBH}{14}
\newcommand{\RNlowIR}{28.6}
\newcommand{\RNlowRob}{10}
\newcommand{\RNmaxTurns}{507}
\newcommand{\RNmedTurns}{35}
\newcommand{\RNmetRhoMax}{0.922}
\newcommand{\RNmetRhoMin}{-0.393}
\newcommand{\RNmetRhoMinName}{delta\_NCD}
\newcommand{\RNmetaA}{18}
\newcommand{\RNmetaB}{19}
\newcommand{\RNminRhoX}{-0.39}
\newcommand{\RNminTurns}{5}
\newcommand{\RNnConv}{202}
\newcommand{\RNnMetrics}{66}
\newcommand{\RNnPairs}{11{,}639}
\newcommand{\RNnPersons}{4}
\newcommand{\RNnTests}{113}
\newcommand{\RNnegTests}{18}
\newcommand{\RNobsSig}{1{,}231}
\newcommand{\RNpersonAShare}{79.1}
\newcommand{\RNpooledRaw}{67}
\newcommand{\RNrawR}{0.048}
\newcommand{\RNrawRho}{0.091}
\newcommand{\RNrepAstr}{28.6}
\newcommand{\RNrepAstrP}{9.5}
\newcommand{\RNrepBstr}{37.1}
\newcommand{\RNrepBstrP}{4.8}
\newcommand{\RNrhoYmal}{0.35}
\newcommand{\RNrhoYrw}{0.16}
\newcommand{\RNrhoYtypx}{0.31}
\newcommand{\RNrobJoint}{35}
\newcommand{\RNrobMO}{35}
\newcommand{\RNrotFac}{1.23}
\newcommand{\RNrotMean}{1{,}003.6}
\newcommand{\RNrotSD}{33.2}
\newcommand{\RNshufFac}{2.42}
\newcommand{\RNshufMean}{509.7}
\newcommand{\RNshufSD}{26.9}
\newcommand{\RNsurveyAbs}{187}
\newcommand{\RNsurveyCand}{443}
\newcommand{\RNsurveyFull}{256}
\newcommand{\RNsurveyNoPool}{48}
\newcommand{\RNsurveyPooled}{208}
\newcommand{\RNtierAconv}{36}
\newcommand{\RNtierApairs}{1{,}672}
\newcommand{\RNtierBconv}{50}
\newcommand{\RNtierBpairs}{2{,}688}
\newcommand{\RNzA}{1.85}
\newcommand{\RNzB}{2.42}
\newcommand{\RRPrefixMetricOnlyFailure}{36.1}
\newcommand{\PEVideoIIDMedian}{1.725}
\newcommand{\PEVideoIIDMin}{0.989}
\newcommand{\PEVideoIIDMax}{4.005}
\newcommand{\PEVideoFisherMedian}{1.606}
\newcommand{\GBootScreenH}{35}
\newcommand{\GBootFSH}{46.2}
\newcommand{\GPublicDailyFS}{0}
\newcommand{\GPublicMeldFS}{0}
\newcommand{\GPublicSwitchFS}{5.3}
\newcommand{\SKnullReps}{5{,}000}
\newcommand{\SKpowReps}{2{,}000}
\newcommand{\SKstudies}{2{,}000}
\newcommand{\SKtol}{0.075}
\newcommand{\SKPoolMaxA}{0.706}
\newcommand{\SKPoolMaxB}{0.661}
\newcommand{\SKJointMax}{0.062}
\newcommand{\SKJointMaxLo}{0.056}
\newcommand{\SKJointMaxHi}{0.069}
\newcommand{\SKJointMin}{0.042}
\newcommand{\SKJointAtLimit}{4}
\newcommand{\SKConsMax}{0.062}
\newcommand{\SKConsMin}{0.000}
\newcommand{\SKConsBelow}{24}
\newcommand{\SKBootMax}{0.063}
\newcommand{\SKBootMin}{0.047}
\newcommand{\SKCRMax}{0.064}
\newcommand{\SKCRMin}{0.046}
\newcommand{\SKFJointMax}{0.520}
\newcommand{\SKFJointOver}{29}
\newcommand{\SKFConsMax}{0.273}
\newcommand{\SKFConsOver}{11}
\newcommand{\SKJointAllMax}{0.063}
\newcommand{\SKPoolIndLo}{0.047}
\newcommand{\SKPoolIndHi}{0.057}
\newcommand{\SKPoolCorpA}{0.201}
\newcommand{\SKPoolCorpB}{0.162}
\newcommand{\SKrHatFiveLo}{0.048}
\newcommand{\SKrHatFiveHi}{0.050}
\newcommand{\SKpowJointFiveMin}{0.282}
\newcommand{\SKpowJointFiveMax}{1.000}
\newcommand{\SKrHatTenLo}{0.099}
\newcommand{\SKrHatTenHi}{0.101}
\newcommand{\SKpowJointTenMin}{0.782}
\newcommand{\SKpowTenAtLeast}{10}
\newcommand{\SKrHatTwentyLo}{0.197}
\newcommand{\SKrHatTwentyHi}{0.201}
\newcommand{\SKfdpPoolMax}{73.2}
\newcommand{\SKfdpPoolMin}{5.8}
\newcommand{\SKfdpJointMax}{3.9}
\newcommand{\SKfdpJointMaxLo}{3.5}
\newcommand{\SKfdpJointMaxHi}{4.2}
\newcommand{\SKtprJointMin}{92.2}
\newcommand{\SKscJointMax}{68.2}
\newcommand{\SKscJointMin}{4.7}
\newcommand{\SKfdpConsMax}{2.7}
\newcommand{\SKfdpBootMax}{5.3}
\newcommand{\SKfdpBootMaxLo}{4.9}
\newcommand{\SKfdpBootMaxHi}{5.7}
\newcommand{\SKfdpCRMax}{7.3}
\newcommand{\SKmcSE}{0.003}
\newcommand{\SKrHatMaxDev}{0.003}
\newcommand{\SKrxHatMid}{0.412}
\newcommand{\SKrxHatHigh}{0.745}
\newcommand{\SKrxHatTop}{0.780}
\newcommand{\SKPoolStrongLo}{0.368}
\newcommand{\SKPoolStrongHi}{0.613}
\newcommand{\SKfdpBootSecond}{4.8}
\newcommand{\SKpowTwentyReps}{24{,}000}
\newcommand{\SKBplusOne}{2{,}001}
\newcommand{\SKmcLimitCount}{301}
\newcommand{\RCownMalLowScr}{14}
\newcommand{\RCownMalLowConf}{7}
\newcommand{\RCownMalLowFail}{50.0}
\newcommand{\RCownMalHighScr}{20}
\newcommand{\RCownMalHighConf}{10}
\newcommand{\RCownAllLowFail}{47.1}
\newcommand{\RCownAllHighFail}{45.2}
\newcommand{\RCnoPsiScr}{64}
\newcommand{\RCnoPsiConf}{35}
\newcommand{\RCnoPsiFail}{45.3}
\newcommand{\RCnoPsiBH}{30}
\newcommand{\RCbhScr}{65}
\newcommand{\RCbhKept}{30}
\newcommand{\RCbhFail}{53.8}
\newcommand{\RCmcSE}{0.007}
\newcommand{\RCmcNearLow}{0.059}
\newcommand{\RCmcNearHigh}{0.062}
\newcommand{\RCpbootEntropy}{0.131}
\newcommand{\RCpbootPsi}{0.115}
\newcommand{\RClouoRobust}{13}
\newcommand{\RClouoOverlap}{8}
\newcommand{\RCpairsMal}{59}
\newcommand{\RCpairsTypx}{35}
\newcommand{\RCpairsRw}{19}
\newcommand{\RCposLateKept}{34}
\newcommand{\RCposBinKept}{32}
\newcommand{\RCwindowR}{-0.127}
\newcommand{\RCwindowLateR}{0.008}
\newcommand{\RCwindowLateN}{7{,}912}
\newcommand{\RCwindowRho}{0.640}
\newcommand{\RCwindowOpenPos}{2{,}719}
\newcommand{\RCwindowOpenN}{2{,}723}
\newcommand{\RCwindowOpenLabel}{15.5}
\newcommand{\RCwindowFullLabel}{29.6}
\newcommand{\RCwithinKept}{6}
\newcommand{\RCwithinHalf}{8}
\newcommand{\RCwithinBelow}{19}
\newcommand{\RCwithinFlip}{2}
\newcommand{\RCwithinMinor}{21}
\newcommand{\RCleadR}{0.186}
\newcommand{\RCleadWithinR}{0.024}
\newcommand{\RChoRobN}{35}
\newcommand{\RChoPoolN}{21}
\newcommand{\RChoSignRobA}{32}
\newcommand{\RChoSignPoolA}{18}
\newcommand{\RChoSignRobB}{31}
\newcommand{\RChoSignPoolB}{14}
\newcommand{\RChoStrictRobA}{10}
\newcommand{\RChoStrictPoolA}{2}
\newcommand{\RCtierALargestPairs}{1{,}313}
\newcommand{\RCtierALargest}{78.5}
\newcommand{\RCbhLastRank}{30}
\newcommand{\RCbhNextRank}{31}
\newcommand{\RCbhLastP}{0.0130}
\newcommand{\RCbhLastThr}{0.0133}
\newcommand{\RCbhNextP}{0.0170}
\newcommand{\RCbhNextThr}{0.0137}
\newcommand{\RCbhSE}{0.004}
\newcommand{\RCrotCells}{10{,}363}
\newcommand{\RCrotPlanned}{22{,}826}
\newcommand{\RCarLowRho}{0.35}
\newcommand{\RCarHighRho}{0.92}
\newcommand{\RCarLow}{3.5}
\newcommand{\RCarHigh}{27.7}
\newcommand{\RCexampleT}{2.13}
\newcommand{\RCnoBothScr}{62}
\newcommand{\RCnoBothConf}{34}
\newcommand{\RCnoBothFail}{45.2}
\newcommand{\RCnoBothBH}{30}
\newcommand{\SDoneStudies}{87}
\newcommand{\SDoneDeclared}{1}
\newcommand{\SDtwoStudies}{220}
\newcommand{\SDtwoTrue}{440}
\newcommand{\SDtwoDeclared}{32}
\newcommand{\SDthreeTrue}{1{,}317}
\newcommand{\SDthreeDeclared}{1{,}307}
\newcommand{\SDrefOneDeclared}{87}
\newcommand{\SDrefTwoDeclared}{440}
\newcommand{\SDtrueMin}{0}
\newcommand{\SDtrueMax}{16}
\newcommand{\SDnullRefLow}{147}
\newcommand{\SDnullRefFewLow}{141}
\newcommand{\SDnullRefHigh}{136}
\newcommand{\SDnullRefFewHigh}{132}
\newcommand{\RBaddB}{17{,}999}
\newcommand{\RBtotalB}{19{,}999}
\newcommand{\RBconfNew}{35}
\newcommand{\RBfailNew}{46.2}
\newcommand{\RBbhNew}{31}
\newcommand{\RBgainOldB}{16}
\newcommand{\RBgainNewB}{106}
\newcommand{\RBgainOldP}{0.0170}
\newcommand{\RBgainNewP}{0.0107}
\newcommand{\RBleadOldP}{0.059}
\newcommand{\RBleadNewP}{0.057}
\newcommand{\RBentNewP}{0.118}
\newcommand{\RBpsiNewP}{0.116}
\newcommand{\RBminP}{0.0001}
\newcommand{\RBedgeRank}{31}
\newcommand{\RBedgeP}{0.0128}
\newcommand{\RBedgeThr}{0.0137}
\newcommand{\RBedgeSE}{0.0011}
\newcommand{\RBbhbhKept}{31}
\newcommand{\RBbhbhFail}{52.3}

\title{The Autocorrelation Blind Spot: Dependence-Aware\\ Inference for Turn-Level Analyses of LLM Conversations}

\author{\name Ferdinand M. Schessl \email ferdinand.schessl@web.de \\ \addr Independent Researcher}

\begin{document}

\maketitle

\begin{abstract}
Turn-level metrics are widely used to evaluate properties of multi-turn human--LLM conversations, from safety and sycophancy to dialogue quality.
However, consecutive turns within a conversation are not statistically independent, and pooled turn-level tests treat them as if they were.
A systematic literature screen (\RNsurveyCand{} candidate records) and a separate full-text audit document pooled turn-level tests without dependence handling; the screen identifies two external studies that handle it statistically, and the audit adds an example using per-person aggregation.
We characterize the autocorrelation structure of \RNnMetrics{} turn-level metrics across \RNnConv{} multi-turn conversations (\RNnPairs{} turn pairs, \RNnPersons{} German-speaking users, one of whom contributes \RNpersonAShare\% of the pairs, 4 LLM platforms): mean per-conversation lag-1 autocorrelation ranges from $\RNmetRhoMin$ (differenced compression distance) to $\RNmetRhoMax$ (cumulative accumulators), and the three annotation label series are themselves autocorrelated (median per-conversation lag-1 of \RNrhoYmal/\RNrhoYtypx/\RNrhoYrw).
Under a dependence-aware procedure that combines the two-series (Bartlett) effective-sample-size correction with a conversation-level block bootstrap, \textbf{\RNirBHJoint\%} of the associations that pass pooled Benjamini--Hochberg screening fail to remain significant (\RNrobJoint{} of \RNbh{} pass at the nominal threshold; BH on the combined values yields \RNbhPfinalJoint{} discoveries, or \RBbhNew{} after a post hoc extension from 2{,}000 to \RBtotalB{} resamples per test that leaves the \RNrobJoint{} unchanged; for one of the \RNbhPfinalJoint, the first difference of a windowed rank metric, the correlation was $\RCwindowR$ over all available turn pairs and $\RCwindowLateR$ after the window had filled, in a post hoc check; \RRPrefixMetricOnlyFailure\% fail under an alternative data construction); on this corpus the bootstrap, which also captures differences between conversations, determines every nominal confirmation.
In simulations with the size of the corpus, the pooled test stays near its nominal level when either series is serially independent and rejects in up to \SKPoolMaxA{} of null cases when both are dependent, while the proposed criterion has a maximum null rejection rate of \SKJointMax{}, below the pre-specified tolerance of \SKtol{}, and a mean false discovery proportion under Benjamini--Hochberg of at most \SKfdpJointMax\% in the simulated studies; with 2{,}000 resamples, Benjamini--Hochberg applied directly to its values over the \RNnTests{} tests declares either no discovery or at least three, and in simulated studies with only one or two true associations these are rarely detected.
In a reanalysis of the pre-registered two-tier hold-out split, with findings classified under the one-series (conservative-bound) form of the criterion, dependence-aware findings replicate more often than pooled-only findings in both tiers (\RNrepAstr\% versus \RNrepAstrP\% in the primary tier).
On three public corpora with per-utterance human labels (DailyDialog, MELD, Switchboard) the design-class ordering of autocorrelation is reproduced and pooled and dependence-aware inference largely agree (failure shares \GPublicDailyFS\%, \GPublicMeldFS\%, and \GPublicSwitchFS\%); on human-annotated counselling dialogues (AnnoMI), resampling whole recordings instead of single utterances widens the intervals of the same 24 correlations by a median factor of \PEVideoIIDMedian{} (range \PEVideoIIDMin--\PEVideoIIDMax).
We provide the inference code, a complete per-pair inventory with validation scripts, definitions of all \RNnMetrics{} metrics, and a publication checklist.
\end{abstract}

\par\ifdim\dimexpr\pagegoal-\pagetotal\relax<6\baselineskip \newpage\fi

\section{Introduction}
\label{sec:intro}

Every turn in a conversation remembers the turn before it.
This fact---obvious to any conversational participant, and well established in discourse analysis, autoregressive language modelling, and dialogue state tracking---is not automatically reflected in the statistical inference of quantitative evaluations of multi-turn LLM interactions.
Turn-level metrics are used for safety classification \citep{inan2023llamaguard, han2024wildguard}, dialogue quality assessment \citep{mehri2020usr, zhang2021dynaeval}, persona consistency tracking, and sycophancy measurement, and we document evaluations in which such turn-level quantities are tested as if \mbox{individual} turns were independent observations (Section~\ref{sec:related}; Appendix~\ref{app:survey}).
They are not.
Lag-1 autocorrelation $\rho_1 = \mathrm{cor}(x_t, x_{t-1})$ quantifies this dependence: $\rho_1 = 0$ indicates no linear dependence between consecutive observations, $\rho_1 \to 1$ indicates strong persistence.
In our corpus of \RNnConv{} human-LLM conversations spanning \RNnPairs{} turn pairs, mean per-conversation lag-1 autocorrelation of the \RNnMetrics{} metrics ranges from $\RNmetRhoMin$ for a differenced compression metric (anti-persistent) to $\RNmetRhoMax$ for a cumulative accumulator.
The consequence for inference is direct: that most persistent metric correlates with a turn-level label at $r = 0.186$ over 11{,}395 turn pairs, which a pooled test declares significant at any conventional level, yet the association is not significant once whole conversations are resampled ($p = 0.059$).
The \emph{label} series carry serial dependence of their own (median per-conversation lag-1 of \RNrhoYmal/\RNrhoYtypx/\RNrhoYrw{} for our three labels), and the null variance of a pooled correlation depends on the dependence structure of \emph{both} series \citep{bartlett1935timecorrelation, pyper1998comparison}.
Turns of one conversation also share its level: conversations differ in their typical metric values and in how often a label occurs, so pooled turns are clustered even where consecutive turns are only weakly correlated.

\paragraph{The scale of the problem.}
We present a systematic empirical characterization of autocorrelation in turn-level conversational metrics.
In our study---an analysis of manipulation detection in LLM conversations using embedding-geometry metrics, with a pre-registered hold-out split---\RNnTests{} metric--label tests yield \RNpooledRaw{} raw-pooled significant associations, \RNbh{} of which pass pooled Benjamini--Hochberg screening \citep{benjamini1995controlling}.
Of these, \RNrobJoint{} remain nominally significant under a dependence-aware procedure combining the joint (two-series) effective-degrees-of-freedom correction \citep{bartlett1935timecorrelation, pyper1998comparison} with a conversation-level block bootstrap \citep{davison1997bootstrap, cameron2015practitioner}: \RNlabMal{} for the \textsc{malicious} label, \RNlabTypx{} for \textsc{typx}, \RNlabRw{} for \textsc{rewiring} (the three annotation labels are defined in Section~\ref{sec:setup}).
The failure share is \RNirBHJoint\% on this screened base (\RNirRawJoint\% on the raw pooled base); Benjamini--Hochberg applied directly to the dependence-aware $p$-values, the primary multiple-testing procedure of this paper, yields \RNbhPfinalJoint{} discoveries (\RBbhNew{} after a post hoc extension to \RBtotalB{} resamples per test; Section~\ref{sec:results}).
The direction of the bias is structural but not universal: \RNnegTests{} of \RNnTests{} tests run on anti-persistent (negatively autocorrelated) metrics, for which, by the two-series formula, pooled inference is conservative for the within-conversation serial component when the label is persistent---and on this corpus no pooled-nonsignificant pair becomes significant under the procedure, so correcting flips no decision in the lenient direction here.

\paragraph{Contributions.}
We make five contributions:
\begin{itemize}[nosep]
    \item[\textbf{C1}] \textbf{Empirical characterization.} We measure per-conversation lag-1 autocorrelation for \RNnMetrics{} turn-level conversational metrics \emph{and} for the annotation label series across \RNnConv{} conversations, and document a spectrum from $\RNmetRhoMin$ to $\RNmetRhoMax$ (Section~\ref{sec:autocorrelation}); the complete per-pair inventory is released.
    \item[\textbf{C2}] \textbf{Dependence-aware correction framework.} We adapt the two-series effective-degrees-of-freedom correction \citep{bartlett1935timecorrelation, pyper1998comparison, bretherton1999effective} and a conversation-level block bootstrap \citep{davison1997bootstrap, cameron2015practitioner} to multi-turn dialogue data, specify Benjamini--Hochberg on the dependence-aware $p$-values as the primary multiple-testing procedure, and evaluate the rejection rates of each component and the detection rates of the combined criterion by simulation, for single tests and for whole studies (Section~\ref{sec:methodology}; Appendix~\ref{app:simulation}).
    \item[\textbf{C3}] \textbf{Hold-out replication.} In a reanalysis of the pre-registered two-tier hold-out split, with findings classified under the one-series (conservative-bound) form of the criterion, dependence-aware findings replicate more often than pooled-only findings in both tiers and under both replication criteria (Section~\ref{sec:holdout}).
    \item[\textbf{C4}] \textbf{Public-corpus evidence.} Standard conversational metrics on three public corpora with per-utterance human labels reproduce the design-class ordering of autocorrelation, with failure shares of \GPublicDailyFS\%, \GPublicMeldFS\%, and \GPublicSwitchFS\%; on human-annotated counselling dialogues, changing only the resampling unit widens the intervals of the same 24 correlations by a median factor of \PEVideoIIDMedian{} (Section~\ref{sec:publicdata}).
    \item[\textbf{C5}] \textbf{Systematic literature survey.} A transparent-query screen of the ACL Anthology 2020--2026 (\RNaclTotal{} entries), arXiv cs.CL, and OpenReview yields \RNsurveyCand{} candidate records; together with a separate full-text audit it documents pooled turn-level tests without dependence handling; the screen identifies two external studies that handle the dependence statistically, and the audit adds an example using per-person aggregation (Section~\ref{sec:related}; Appendix~\ref{app:survey}).
\end{itemize}

\paragraph{Scope and running example.}
Our running example is manipulation detection in human-LLM conversations, where we correlate embedding-geometry metrics with turn-level annotations of manipulative behavior.
The corpus comes from four users, one of whom contributes \RNpersonAShare\% of the turn pairs; it is an application case, not a population sample (Section~\ref{sec:users}).
The methodological problem we identify is a deductive consequence of conversational architecture: in any multi-turn conversation, the LLM's response at turn~$t$ is conditioned on the full prefix $(x_1, \ldots, x_{t-1})$; the user's subsequent message is shaped by this response; and the LLM's next output depends on both.
This causal chain lets turn-level metrics retain serial dependence, and how much they retain depends on how a metric is constructed.
The specific failure share is corpus-dependent (Section~\ref{sec:publicdata}), and the effect of serial dependence in both series on pooled tests is studied under controlled conditions by simulation (Appendix~\ref{app:simulation}).

\section{Background and Related Work}
\label{sec:related}

\subsection{Turn-Level Evaluation in Conversational AI}
\label{sec:related-turnlevel}

The dominant paradigm for evaluating multi-turn LLM interactions computes per-turn scores and aggregates them.
Safety classifiers such as LlamaGuard \citep{inan2023llamaguard} and WildGuard \citep{han2024wildguard} assign risk scores to individual messages.
Dialogue quality metrics like USR \citep{mehri2020usr} and DynaEval \citep{zhang2021dynaeval} evaluate response appropriateness at the turn level.
Engagement and coherence trackers measure properties of successive responses.
When such scores are analyzed statistically---with correlations, $t$-tests, or regression coefficients---turns are typically treated as rows in a flat table.

This approach inherits a structural problem: consecutive turns within a conversation are not independent (Section~\ref{sec:intro}).
Metrics computed on these turns can retain this serial dependence.
When the series are positively persistent, the result is \emph{pseudoreplication} \citep{hurlbert1984pseudoreplication}: the pooled turns carry less information---sometimes much less---than their number suggests.

\subsection{Dependence-Aware Inference in Statistics and Other Fields}
\label{sec:related-autocorrelation}

The statistical consequences of temporal autocorrelation have been characterized for nearly a century, and the corrections we transfer are standard in several fields.

\paragraph{Statistics: the two-series correction.}
\citet{bartlett1935timecorrelation} derived the variance of the sample correlation between two autocorrelated series: to first order, $\operatorname{Var}(\hat r) \approx (1 + \rho_x \rho_y)/\bigl(n (1 - \rho_x \rho_y)\bigr)$ under AR(1) dependence in both series, so the effective sample size depends on the \emph{product} of the two lag-1 autocorrelations.
\citet{pyper1998comparison} compare estimators of this correction in ecological time series, and \citet{bretherton1999effective} give the corresponding effective degrees of freedom for climate fields.
Section~\ref{sec:chelton} adopts the two-series form as the analytic correction and retains the one-series special case as a labeled conservative bound.

\paragraph{Climate science.}
\citet{chelton1983effects} discusses sampling errors in statistical estimation in the oceanographic literature; effective-degrees-of-freedom corrections for autocorrelated series are standard practice in physical oceanography and climate science \citep{mudelsee2010climate}.

\paragraph{Neuroscience.}
\citet{eklund2016cluster} showed that spatial autocorrelation in fMRI data inflates cluster-level false positive rates far above the nominal 5\%, prompting field-wide reanalysis and new correction standards.

\paragraph{Psychotherapy process research.}
\citet{tasca2009multilevel} argued that session-level observations in psychotherapy are nested within patients and serially dependent within treatment, requiring multilevel models with autoregressive error structures.
Their recommendation---that ignoring this nesting inflates Type I error---applies with equal force to turn-level observations nested within conversations.

\paragraph{Econometrics.}
\citet{newey1987simple} developed heteroskedasticity and autocorrelation consistent (HAC) estimators for regression with time series data; cluster-robust standard errors \citep{cameron2015practitioner} extend this logic to grouped data (analogous to turns grouped within conversations).

\paragraph{In NLP: known in principle, with documented omissions in practice.}
NLP research is well aware that consecutive turns are sequentially dependent: this is the foundation of autoregressive language modelling, discourse analysis, and dialogue state tracking, and remains an active area of study in multi-turn evaluation \citep{luzdearaujo2026persistent}.
The gap we document is narrower and strictly statistical: evaluations that report pooled turn-level tests without dependence-aware inference.
The general point has been made for NLP test sets, whose items are rarely independent \citep{dror2018hitchhiker}, and for language-model evaluations, where clustered standard errors are recommended when questions are drawn in related groups \citep{miller2024errorbars}; turns within conversations are such a grouping.
We ran a systematic screen with transparent queries over the ACL Anthology 2020--2026 (\RNaclTotal{} entries), arXiv cs.CL, and OpenReview (ICLR/NeurIPS), yielding \RNsurveyCand{} candidate records, of which \RNsurveyFull{} were screened on full text; an automated proximity screen flags \RNsurveyPooled{} of these for pooled turn-level inference.\footnote{Protocol, queries, screening rules, and the complete record table are in Appendix~\ref{app:survey} and the supplement. The flags select passages for inspection; they are not hand-verified classifications, and the screen does not estimate how common dependence handling is in the field.}
Among the flagged records, hand inspection of all hits for dependence-handling vocabulary identifies two external studies that handle the dependence statistically: survival analysis with conversation-level cluster-robust standard errors \citep{li2025timetoinconsistency}, and per-trajectory aggregation with the serial dependence of the underlying drift measure reported explicitly \citep{wu2026sellme}.
A separate 30-entry audit of the available full-text assessments, including one entry whose bibliographic identity remains uncertain, supplies directly checked cases: in its documented classification, which includes borderline cases, five apply no dependence handling---among them a test over 15{,}345 pooled responses---four handle it partially, and one uses per-person aggregation (Appendix~\ref{app:survey-list}).
This is not a criticism of individual papers, and it is not an estimate of how common either practice is; the documented omissions motivate the reporting we recommend, and the positive examples show that the tools are compatible with current evaluation practice.

\subsection{Cluster-Robust Inference}
\label{sec:related-cluster}

The statistical tools for handling autocorrelation in grouped data exist and are well understood.
Block and cluster bootstrap methods \citep{politis1994stationary, davison1997bootstrap} resample entire blocks (in our case, entire conversations) rather than individual observations, preserving the within-block dependence structure.
Cluster-robust standard errors \citep{cameron2015practitioner} adjust variance estimates by treating each cluster (conversation) as the independent sampling unit.
With few clusters or clusters of very unequal size these methods can over-reject, and bootstrap refinements and small-sample corrections have been developed for such cases \citep{cameron2008bootstrap, pustejovsky2018small, mackinnon2023cluster}; the simulation of Appendix~\ref{app:simulation} uses the empirical conversation lengths, and the level of users, with four clusters of which one holds \RNpersonAShare\% of the turn pairs, is treated as a sensitivity analysis only (Section~\ref{sec:users}).
Random-effects models with by-item random slopes \citep{barr2013random} offer a complementary approach for crossed random effects.

Our contribution is not the invention of new statistical methods.
It is the \emph{adaptation} of these methods for the specific dependence structure of conversational data---where turns are nested within conversations, conversations vary in length by two orders of magnitude, both metric and label series are serially dependent, and the autocorrelation structure differs widely across metric types---and the \emph{empirical measurement} of how many pooled findings fail under dependence-aware inference in a real evaluation setting.

\section{The Autocorrelation Problem in Conversational Data}
\label{sec:autocorrelation}

Turn-level metrics extracted from multi-turn conversations are not independent observations.
Each turn is a response to the preceding one, and many commonly used metrics---cumulative scores, rolling averages, trajectory summaries---carry forward information from earlier turns by construction.
When the persistence of a metric and of a label align, pooled tests can overstate the evidence behind a correlation computed over turn-level data relative to what the nominal sample size suggests.
In this section, we characterize the severity of this problem empirically across our corpus and demonstrate its consequences for hypothesis testing.

\subsection{Empirical Characterization}
\label{sec:empirical-autocorrelation}

We compute the lag-1 autocorrelation $\rho_1$ for each turn-level metric within each of the \RNnConv{} conversations, then average across conversations to obtain $\bar{\rho}_1$; Appendix~\ref{app:metrics} states the estimator and its conventions for short and constant sequences.

Table~\ref{tab:autocorrelation} presents these values for all thirteen metric families of the battery, ordered by decreasing autocorrelation.
The range is wide: from $\bar{\rho}_1 = 0.910$ for cumulative accumulators to values near zero for timestamp/recovery and directional-consistency metrics, with individual metrics spanning $\RNmetRhoMin$ (differenced compression distance, anti-persistent) to $\RNmetRhoMax$ (cumulative accumulator).

\begin{table}[t]
\centering
\small
\begin{tabular}{@{}lcc@{}}
\toprule
\textbf{Metric family} & $\bar{\rho}_1$ & \textbf{Metrics} \\
\midrule
Cumulative accum. & 0.910 & 2 \\
Normalized cumul. & 0.763 & 1 \\
Rolling aggr. (mixed W) & 0.713 & 2 \\
Weighted variants & 0.606 & 3 \\
Frame distance & 0.411 & 4 \\
Lexical/structural & 0.378 & 9 \\
Thermodynamic & 0.367 & 13 \\
Interaction & 0.256 & 5 \\
Embedding velocity & 0.169 & 12 \\
Impulse/indicator & 0.115 & 7 \\
Compression (NCD) & 0.106 & 1 \\
Timestamp/recovery & 0.051 & 6 \\
Directional consist. & $-$0.014 & 1 \\
\bottomrule
\end{tabular}
\caption{%
  Mean lag-1 autocorrelation $\bar{\rho}_1$ by metric family (per-conversation values averaged across the \RNnConv{} conversations, then across the distinct metrics of each family) and number of distinct metrics per family.
  Each metric enters once, with the observations of its first test in the battery.
  Per-metric values for all \RNnMetrics{} metrics are in the released inventory; Figure~\ref{fig:autocorrelation} visualizes the family means.%
}
\label{tab:autocorrelation}
\end{table}

\begin{figure}[t]
\centering
\includegraphics[width=0.62\linewidth]{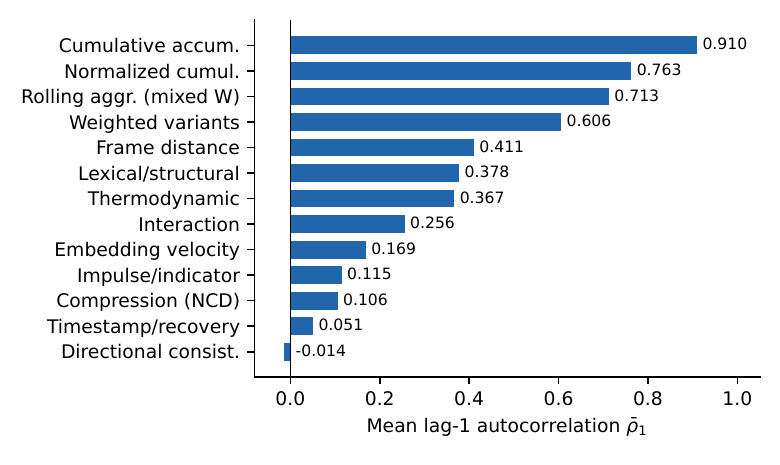}
\caption{Mean lag-1 autocorrelation across all thirteen metric families. Accumulation and overlapping windows carry information across turns; per-turn quantities and their first differences carry little, and individual differenced metrics are anti-persistent, while the first difference of a windowed quantity can remain persistent ($\bar\rho_1 = \RCwindowRho$ for \texttt{delta\_NC2}; Section~\ref{sec:results}).}
\label{fig:autocorrelation}
\end{figure}

Several patterns are notable.
First, the ordering is not arbitrary: the families at the top are built by accumulation or overlapping windows, which carry earlier turns forward.
Cumulative accumulators (e.g., running budget scores, cumulative information measures) carry all previous information forward; their autocorrelation approaches that of a random walk ($\rho \to 1$ for large~$n$).
Rolling aggregates behave similarly because consecutive windows overlap by $W{-}1$ observations.
At the other extreme, per-turn quantities such as embedding velocity ($\|\Delta\mathbf{e}_t\| = \|\mathbf{e}_t - \mathbf{e}_{t-1}\|$) measure local change with minimal carryover, and first differences of per-turn quantities can be anti-persistent (down to $\bar\rho_1 = \RNmetRhoMin$ for \RNmetRhoMinName).
Second, the families are thematic groupings that mix constructions, so the variation \textit{within} a family can be substantial, and a small family mean can result from persistent and anti-persistent members cancelling; Appendix~\ref{app:metrics} gives the construction of each metric and the released inventory its autocorrelation.

\subsection{Consequences for Hypothesis Testing}
\label{sec:consequences}

To make the consequences concrete, we present a worked example.
Consider a metric with $\bar{\rho}_x = 0.8$, measured over a conversation of $n = 200$ turn pairs, correlated with a label series of moderate persistence ($\bar\rho_y = 0.33$).
A pooled point-biserial correlation of $r = 0.15$ with the standard test statistic
\begin{equation}
t = r \sqrt{\frac{n-2}{1-r^2}}
\label{eq:t-standard}
\end{equation}
gives $t = \RCexampleT$ with $\text{df} = 198$, corresponding to $p = 0.034$---significant at $\alpha = 0.05$ by conventional standards.

Applying the joint correction (Eq.~\ref{eq:chelton-joint} below) with $\bar\rho_x \bar\rho_y = 0.264$ yields
\begin{equation}
n_{\mathrm{eff}} = 200 \times \frac{1 - 0.264}{1 + 0.264} \approx 116.5,
\qquad t_{\mathrm{corr}} = 0.15 \times \sqrt{\frac{114.5}{0.9775}} = 1.62,
\end{equation}
with $p_{\mathrm{corr}} = 0.11$: the finding does not survive.
Under the conservative one-series bound ($\rho_y \equiv 1$, Eq.~\ref{eq:chelton}), $n_{\mathrm{eff}} = 200 \times 0.2/1.8 \approx 22.2$ and $p_{\mathrm{corr}} \approx 0.50$.
The \textit{effect size} is unchanged ($r = 0.15$), but the \textit{evidence} for it is far weaker than the pooled analysis suggests: between roughly two-fifths and nine-tenths of the nominal degrees of freedom are absorbed by the serial dependence of the two series, with the exact share set by the product $\bar\rho_x\bar\rho_y$.

The example isolates the serial component of the dependence; within the pooled BH screen of our corpus, the conversation-level bootstrap determines nominal confirmation (Section~\ref{sec:results}), and it also preserves the differences between conversations introduced next.

\subsection{Why Conversations Are Especially Vulnerable}
\label{sec:why-vulnerable}

Serial dependence is a known issue in time-series analysis \citep{bartlett1935timecorrelation, chelton1983effects, von1941statistical}.
The evaluations documented in Section~\ref{sec:related} show that the standard corrections are not always applied to conversation data.
Five features of conversational data make the problem especially severe:

\paragraph{Inherent sequential dependence.}
Conversations are \textit{causally} sequential: each turn is a direct response to the previous one.
Metric design can change the observed autocorrelation, but it cannot turn the turns of one conversation into independent sampling units.

\paragraph{Variable conversation length.}
Conversations in our corpus range from \RNminTurns{} to \RNmaxTurns{} turn pairs (median \RNmedTurns).
When observations are pooled across conversations, long conversations dominate the analysis despite contributing relatively few independent observations.
A conversation with 500 turn pairs and $\rho = 0.94$ contributes 500 nominal observations but only ${\sim}15$ effective ones under the conservative bound.
Pooling treats all 500 equally, which overstates the precision of the estimate.

\paragraph{Both series are dependent.}
Turn-level label series---whether human or model-generated---are persistent whenever the labeled construct persists (a manipulative episode spans many turns).
In our corpus the three label series have median per-conversation lag-1 autocorrelation of \RNrhoYmal, \RNrhoYtypx, and \RNrhoYrw{} (Appendix~\ref{app:annotation}), and it is the \emph{product} of metric and label dependence that enters the null variance (Section~\ref{sec:chelton}); in the simulation study, pooled testing stays near its nominal level when either series is serially independent and exceeds it increasingly as both become dependent (Appendix~\ref{app:simulation}).

\paragraph{Conversations differ from each other.}
Turns of one conversation share more than their order: conversations differ in their typical metric level and in how often a label occurs.
We describe this by the share of a variable's total variation that lies between conversation means, $V_{\mathrm{between}}(x)=\sum_c n_c(\bar x_c-\bar x)^2/\sum_c\sum_{t}(x_{ct}-\bar x)^2$.
In our corpus this share is 40.3\% for the \textsc{malicious} label, 56.2\% and 60.1\% for two cumulative metrics, and 12.7\% for embedding velocity.
Serial dependence alone produces part of such a share: for a stationary AR(1) series with the same mean in every conversation and the conversation lengths of this corpus, the ratio of the expected between-conversation to the expected total sum of squares is \RCarLow\% at a lag-1 value of \RCarLowRho{} and \RCarHigh\% at \RCarHighRho.
The observed label share of 40.3\% exceeds this illustrative reference; the comparison does not identify a conversation-level variance component.
For the lead example of Section~\ref{sec:intro}, \textsc{malicious} and \texttt{Toughness}, the pooled correlation is \RCleadR{} and the correlation after centering both series within conversations is \RCleadWithinR{} (a post hoc description; Section~\ref{sec:results}).
Lag-1 autocorrelation within conversations does not capture this component of the dependence; treating the conversation as the unit of resampling does (Section~\ref{sec:bootstrap}).

\paragraph{Loss of power in short conversations.}
For a conversation of 30 turn pairs with $\rho_x = 0.8$ and $\rho_y = 0.33$, the joint effective sample size is $30 \times 0.736/1.264 \approx 17$.
Such dependence can reduce power within short conversations; evidence can accumulate across independent conversations.

\section{Correction Methodology}
\label{sec:methodology}

We propose a correction protocol built from two complementary components---an analytic correction based on effective degrees of freedom that accounts for the dependence of \emph{both} series, and a nonparametric correction based on cluster resampling---and we specify the multiple-testing procedure operating on the corrected $p$-values.
Each component is evaluated by simulation under controlled metric and label dependence (Appendix~\ref{app:simulation}).

\subsection{Effective Sample Size for Two Autocorrelated Series}
\label{sec:chelton}

For the correlation of two mutually independent AR(1) series, the large-sample variance of $\hat r$ gives the effective sample size \citep{bartlett1935timecorrelation, pyper1998comparison, bretherton1999effective}
\begin{equation}
n_{\mathrm{eff}} = n \cdot \frac{1 - \bar{\rho}_x \bar{\rho}_y}{1 + \bar{\rho}_x \bar{\rho}_y},
\label{eq:chelton-joint}
\end{equation}
where $\bar{\rho}_x$ and $\bar{\rho}_y$ are the mean lag-1 autocorrelations of the metric and the label series.
For comparison, setting $\rho_y = 1$ in this expression gives the one-series form
\begin{equation}
n_{\mathrm{eff}}^{(\mathrm{cons})} = n \cdot \frac{1 - \bar{\rho}_x}{1 + \bar{\rho}_x},
\label{eq:chelton}
\end{equation}
an algebraic limiting case, not a stationary label process; for nonnegative metric autocorrelation it is the smallest member of the family, and we retain it under the label ``conservative bound'', which describes this ordering and is not a demonstrated error control.

In practice, we estimate $\bar{\rho}_x$ by computing the lag-1 autocorrelation of the metric series within each conversation and averaging across conversations, and $\bar{\rho}_y$ analogously over the conversations in which the label varies (a constant label series has no defined autocorrelation); both are computed on the turn pairs available for the test.
Treating constant-label conversations as $\rho_y = 0$ instead leaves the set of robust pairs unchanged; both variants are in the released inventory.
We impose the floor
\begin{equation}
n_{\mathrm{eff}} \geq k,
\label{eq:lower-bound}
\end{equation}
with $k$ the number of conversations, so that each conversation counts as at least one observation; the floor is a convention of the procedure, not a theoretical bound.
For anti-persistent metrics ($\bar\rho_x < 0$; \RNnegTests{} of \RNnTests{} tests, down to $\bar\rho_x = \RNminRhoX$) the product $\bar\rho_x\bar\rho_y$ is negative when the label is persistent, and Eq.~\ref{eq:chelton-joint} yields $n_{\mathrm{eff}} > n$: the corrected $p$-value is smaller than the pooled one, and the final criterion below still requires the bootstrap to agree, so $n_{\mathrm{eff}} > n$ never rescues a pair on its own.

The corrected $p$-value uses the test statistic of Eq.~\ref{eq:t-standard} with $n_{\mathrm{eff}}$ in place of $n$, $t_{\mathrm{corr}} = r\sqrt{(n_{\mathrm{eff}}-2)/(1-r^2)}$, and $\text{df} = n_{\mathrm{eff}} - 2$ degrees of freedom of the $t$-distribution:
\begin{equation}
p_{\mathrm{Chelton}} = 2 \cdot P\!\left(T > |t_{\mathrm{corr}}|\right), \quad T \sim t(n_{\mathrm{eff}} - 2).
\label{eq:p-chelton}
\end{equation}
The subscript is a label kept from the released materials; it does not attribute the formula.

\paragraph{Limitations.}
The correction assumes stationarity and geometrically decaying (AR(1)) dependence in both series.
Cumulative metrics exhibit higher-order autocorrelation, and averages of within-conversation estimates do not describe the differences between conversations (Section~\ref{sec:why-vulnerable}).
The analytic correction can therefore be insufficient on its own, motivating the complementary nonparametric approach below.
In the simulation study, which contains neither cumulative metrics nor conversation-specific effects, the analytic corrections alone still exceed the nominal level in many cells with strong dependence of both series (up to \SKFJointMax{} for the joint form).
In every null cell, the estimates of both lag-1 correlations used in the correction average below their population targets.
The combined criterion stays within the pre-specified tolerance (Appendix~\ref{app:simulation}).

\subsection{Cluster-Robust Block Bootstrap}
\label{sec:bootstrap}

The block bootstrap \citep{efron1979bootstrap, davison1997bootstrap}, applied with whole conversations as blocks (a cluster bootstrap), provides a nonparametric alternative that needs no model of the dependence within conversations.
The key insight is that resampling \textit{entire conversations} as atomic units preserves the within-conversation dependence of both series, and the differences between conversations, without the need to model either explicitly.

Algorithm~\ref{alg:bootstrap} specifies the procedure.
We resample $k$ conversations with replacement from the pool of $k$ conversations, concatenate all turns from the sampled conversations into a pooled dataset, and compute the point-biserial correlation $r_{b'}$ on this bootstrap sample.
After $B = 2{,}000$ iterations, let $b$ be the number of bootstrap correlations whose sign differs from that of the observed correlation $r$ ($r_{b'} \leq 0$ if $r > 0$, $r_{b'} \geq 0$ otherwise).
The two-sided $p$-value, an unstudentized tail-area value with the add-one convention for resampling tails, is
\begin{equation}
p_{\mathrm{boot}} = \min\!\Bigl\{1,\; \frac{2\,(b+1)}{B+1}\Bigr\}.
\label{eq:p-boot}
\end{equation}

\begin{algorithm}[t]
\caption{Cluster-Robust Block Bootstrap}
\label{alg:bootstrap}
\begin{algorithmic}[1]
\REQUIRE Conversations $C_1, \ldots, C_k$ with turns $\{(\ell_i^{(j)}, m_i^{(j)})\}_{i=1}^{n_j}$; observed correlation $r$
\REQUIRE Number of bootstrap iterations $B = 2{,}000$
\ENSURE Two-sided $p$-value $p_{\mathrm{boot}}$
\FOR{$b' = 1$ \TO $B$}
    \STATE Sample $k$ conversations with replacement: $\{C_{s_1}^*, \ldots, C_{s_k}^*\}$
    \STATE Pool all turns, keeping repeated draws: $\mathcal{D}_{b'}^* \leftarrow \biguplus_{j=1}^{k} \{(\ell_i^{(s_j)}, m_i^{(s_j)})\}$
    \STATE Compute $r_{b'} \leftarrow \text{pointbiserialr}(\boldsymbol{\ell}_{b'}^*, \boldsymbol{m}_{b'}^*)$; set $r_{b'} \leftarrow 0$ if either variable is constant
\ENDFOR
\STATE $b \leftarrow \#\{b' : r_{b'} \leq 0\}$ if $r > 0$, else $\#\{b' : r_{b'} \geq 0\}$
\RETURN $p_{\mathrm{boot}} \leftarrow \min\{1,\, 2(b+1)/(B+1)\}$
\end{algorithmic}
\end{algorithm}

Here, $\ell_i^{(j)} \in \{0, 1\}$ denotes the binary label and $m_i^{(j)} \in \mathbb{R}$ denotes the metric value for turn $i$ in conversation $j$.
The point-biserial correlation $r$ is algebraically equivalent to the Pearson correlation between a binary and a continuous variable.

\paragraph{Why conversations, not turns?}
Resampling individual turns would destroy the sequential structure and produce samples with near-zero autocorrelation---exactly what the procedure must avoid.
If a metric's significance is driven by patterns within a few conversations and not by evidence that accumulates across conversations, the bootstrap will produce a wide distribution of $r_{b'}$ values that spans zero, yielding a high $p_{\mathrm{boot}}$.
In the simulation study, the bootstrap alone has null rejection rates between \SKBootMin{} and \SKBootMax{} across all null cells, including the cells where the analytic corrections alone reject far too often (Appendix~\ref{app:simulation}); it is therefore a mandatory component of the criterion.
The statistic remains the pooled correlation: resampling whole conversations changes the uncertainty attached to it, not the quantity being estimated, and it treats conversations as the independent sampling unit (Section~\ref{sec:users} examines the level of users).
This marginal association includes within- and between-conversation covariance and weights observations equally, so longer conversations contribute more; it is not a within-conversation effect estimate.

\subsection{Testing Protocol and Multiple-Testing Procedure}
\label{sec:protocol}

For each metric--label pair we compute the pooled $p$-value, $p_{\mathrm{Chelton}}$ (Eq.~\ref{eq:p-chelton}, joint form), and $p_{\mathrm{boot}}$ (Eq.~\ref{eq:p-boot}), and define the dependence-aware $p$-value
\begin{equation}
p_{\mathrm{final}} = \max(p_{\mathrm{Chelton}},\; p_{\mathrm{boot}}),
\label{eq:robust-criterion}
\end{equation}
requiring the analytic and the nonparametric correction to agree.
The maximum of two $p$-values is valid whenever at least one of them is.

\paragraph{Primary procedure: BH on the dependence-aware $p$-values.}
The primary multiple-testing procedure of this paper applies Benjamini--Hochberg at $q = 0.05$ directly to the $p_{\mathrm{final}}$ values of all \RNnTests{} tests.
The usual BH guarantee presupposes valid null $p$-values and independence or a suitable form of positive dependence across tests; neither is established here by proof.
On our battery it declares \RNbhPfinalJoint{} discoveries, a strict subset of the \RNrobJoint{} pairs below.
A post hoc extension to \RBtotalB{} resamples per test gives \RBbhNew, again a subset of the same \RNrobJoint{} (Section~\ref{sec:results}).
In the study-level simulation its observed mean false discovery proportion is at most \SKfdpJointMax\% across ten conditions, including studies without and with few true associations.
With $B=2{,}000$, the bootstrap resolution permits either zero or at least three BH discoveries, so a study with only one or two true associations can report them only together with false discoveries; in the stored simulated studies of this kind they are rarely detected (Appendix~\ref{app:simulation}).
The present analysis does not resolve an isolated rank-one or rank-two discovery.
The global-null comparison with the continuous cluster-robust reference therefore applies to this resampling budget (Appendix~\ref{app:simulation}).
In the extension to \RBtotalB{} resamples the smallest attainable value, \RBminP, lies below the first BH threshold; error rates and power at that budget were not simulated.

\paragraph{Reported alongside: screen-then-confirm.}
A two-stage variant---BH screening on the pooled $p$-values, then per-pair confirmation by $p_{\mathrm{final}} < \alpha$---is reported alongside, because its screening stage is the natural place to define the failure share (Section~\ref{sec:inflation-rate}).
Its screening stage uses pooled $p$-values and its confirmation is per pair, so it carries no FDR guarantee; in the study-level simulation its mean false discovery proportion ranges from \SKscJointMin\% to \SKscJointMax\% and exceeds 5\% in eight of ten conditions, so the pairs it confirms are not a multiplicity-controlled set.
The released code implements both procedures.

\subsection{The Failure Share as Diagnostic}
\label{sec:inflation-rate}

We define the \textit{failure share} ($\IR$) as the proportion of pooled-significant findings that fail to remain significant under the specified dependence-aware procedure:
\begin{equation}
\IR = \frac{n_{\mathrm{pooled\text{-}sig}} - n_{\mathrm{robust}}}{n_{\mathrm{pooled\text{-}sig}}}.
\label{eq:inflation-rate}
\end{equation}
Here $n_{\mathrm{robust}}$ counts the tests in the pooled screen that also pass the specified dependence-aware procedure; an empty screen leaves the share undefined.
This is a disagreement rate between testing procedures, not a direct estimate of the fraction of false associations: a pair can fail the corrected test because it was a false positive of pooled testing \emph{or} because the corrected test has less power.
The simulation illustrates both components under known truth: at an effect of $r = 0.10$ the combined criterion detects the association in at least 95\% of the replications in \SKpowTenAtLeast{} of 12 simulated conditions, and less often for weak effects under strong dependence of both series (Appendix~\ref{app:simulation}).
We use the neutral operational phrase ``fails to remain significant under the specified dependence-aware procedure'' throughout, and reserve false-positive language for the simulation, where ground truth is known.

\paragraph{Corpus-level result.}
In our corpus (\RNnConv{} conversations, \RNnPairs{} turn pairs, \RNnTests{} tests), \RNpooledRaw{} pairs are raw-pooled significant and \RNbh{} pass pooled BH screening; \RNrobJoint{} pairs remain nominally significant under the dependence-aware procedure (we call them \emph{robust}, and pairs that pass only the pooled test \emph{pooled-only}), and every robust pair lies inside the BH-screened set.
The failure share is therefore
\begin{equation}
\IR = \frac{\RNbh - \RNrobJoint}{\RNbh} \approx \RNirBHJoint\%
\end{equation}
on the screened base and \RNirRawJoint\% on the raw pooled base.
Comparing BH on the pooled and on the dependence-aware $p$-values instead, \RCbhKept{} of the \RCbhScr{} pooled discoveries remain (\RCbhFail\% fail; \RBbhbhKept{} and \RBbhbhFail\% with \RBtotalB{} resamples per test); this multiplicity-matched comparison is distinct from the screen-then-confirm diagnostic of \RNirBHJoint\%.

\begin{figure}[t]
\centering
\includegraphics[width=0.8\linewidth]{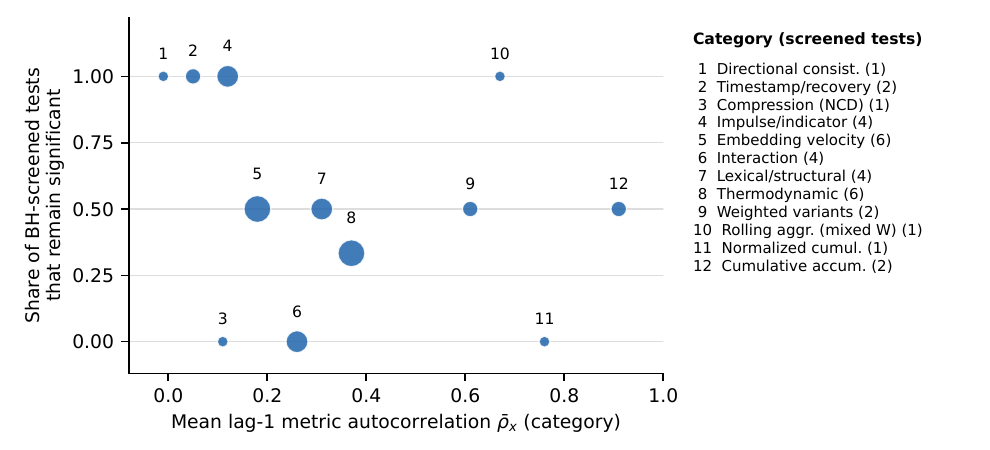}
\caption{Share of BH-screened \textsc{malicious} tests that remain significant under the dependence-aware procedure, per metric category, against the category's mean lag-1 metric autocorrelation. Numbers identify the categories; point area is proportional to the number of screened tests (in parentheses). The values are those of Table~\ref{tab:main}; the Frame distance category has no screened test and is not shown.}
\label{fig:survival}
\end{figure}

\paragraph{Decomposition by autocorrelation.}
The failure share varies across categories from 0\% to 100\% (Table~\ref{tab:main}; Figure~\ref{fig:survival}).
On the \textsc{malicious} label, the families whose mean autocorrelation in Table~\ref{tab:autocorrelation} is at most 0.25 retain \RNlowRob{} of \RNlowBH{} screened pairs (failure share \RNlowIR\%), the more persistent families \RNhiRob{} of \RNhiBH{} (\RNhiIR\%).
This contrast between families is descriptive and depends on the threshold (assigning the Interaction family, mean 0.256, to the other group gives 44.4\% and 56.3\%) and on the grouping: split by the autocorrelation of their own metric at 0.25, the same tests give \RCownMalLowConf{} of \RCownMalLowScr{} and \RCownMalHighConf{} of \RCownMalHighScr{} surviving pairs, a failure share of \RCownMalLowFail\% in both groups (\RCownAllLowFail\% and \RCownAllHighFail\% among the \RNbh{} screened tests of all three labels).
The contrast between families therefore does not establish a corresponding ordering of individual tests by metric persistence.

The failure share reports a disagreement between the two specified procedures; we recommend reporting it together with the lag-1 autocorrelations of both series (Section~\ref{sec:recommendations}).

\section{Empirical Demonstration}
\label{sec:empirical}

We now apply the correction protocol from Section~\ref{sec:methodology} to a
multi-turn manipulation detection task.  The goal is not to validate any
particular metric---the metrics serve as a running example---but to quantify
\emph{how many} pooled findings fail to remain significant once the dependence is respected.

\subsection{Experimental Setup}
\label{sec:setup}

\paragraph{Corpus.}
We analyze \RNnConv{} multi-turn conversations between four German-speaking
users and four LLM platforms (ChatGPT (GPT-4/GPT-4o), DeepSeek, Gemini, and
Claude), totalling \RNnPairs{} aligned turn pairs.  Conversations range from
\RNminTurns{} to \RNmaxTurns{} turn pairs (median \RNmedTurns).  They cover
everyday topics (cooking, travel, study assistance) as well as adversarial
interactions during extended use.
Table~\ref{tab:users} reports the corpus composition per user: the corpus is
strongly imbalanced, with the largest user contributing \RNpersonAShare\% of
turn pairs; Section~\ref{sec:users} quantifies the consequences.

\begin{table}[t]
\centering
\small
\begin{tabular}{@{}llll@{}}
\toprule
Person & Chats & Pairs & Pair share (percent) \\
\midrule
Person A & 140 & 9{,}212 & 79.1 \\
Person B & 34 & 1{,}173 & 10.1 \\
Person C & 22 & 737 & 6.3 \\
Person D & 6 & 517 & 4.4 \\
\bottomrule
\end{tabular}
\caption{Corpus composition per user (anonymized).  All user-level analyses
(leave-one-user-out, user-level cluster bootstrap; Section~\ref{sec:users})
use this partition.  Percentages may not sum to 100 because of rounding.}
\label{tab:users}
\end{table}

\paragraph{Task and Labels.}
Each turn pair carries a binary \textsc{malicious} label
($\textsc{mal} \in \{0,1\}$) indicating whether the assistant's
response exhibits manipulative characteristics, as determined by a
structured annotation pipeline,\footnote{Annotation uses a structured
LLM-as-judge protocol, with a qualitative review by a computational
linguist; see Appendix~\ref{app:annotation}.  The label series are themselves
serially dependent (median per-conversation lag-1 of
\RNrhoYmal/\RNrhoYtypx/\RNrhoYrw{} for the three labels), which enters the
corrected inference through Eq.~\ref{eq:chelton-joint} and is preserved by
conversation-level resampling.}
alongside two further binary labels used in the cross-label analyses,
\textsc{typx} (emergent co-construction) and \textsc{rewiring} (semantic
redefinition of established terms).
The judge receives a summary of the whole conversation and, with each batch
of turns, the preceding five annotation rows; label persistence can therefore
reflect both the persistence of the annotated behavior and this shared context
(Appendix~\ref{app:annotation}).

\paragraph{Pre-registration.}
Two internal pre-registration documents were filed before hold-out
annotation: the original design (dated 2026-03-10) and a two-tier revision
(dated 2026-03-18) specifying a hold-out of 50 conversations split into
Tier~A (\RNtierAconv{} strictly unseen conversations) and Tier~B (all 50,
including 14 conversations touched by a processing pilot), six directional
hypotheses with a Holm criterion, and a meta-level count criterion.
The remaining \RNexplConv{} conversations form the exploratory set.
Both documents are included in the supplement in anonymized form.
The analyses below retain the pre-registered split, hypotheses, and decision
rules; the metrics follow the data construction described below, which was
chosen after the first hold-out results were known, so the hold-out results are
a reanalysis and not an untouched confirmatory test.

\paragraph{Metrics.}
We compute \RNnMetrics{} turn-level metrics, grouped into thirteen
families (Table~\ref{tab:autocorrelation}) that range from per-turn embedding
velocities and first differences to rolling-window aggregates and cumulative
accumulators.  Every metric is defined in
Appendix~\ref{app:metrics} (formula, inputs, windows, design family).
A small number of engine parameter \emph{values} are withheld for
responsible-disclosure reasons and are marked as such there; all reported
statistics are computed from the stored metric values.
We retain the \RNnTests{} metric--label pairs of the original exploratory
analysis (\RCpairsMal{} \textsc{malicious}, \RCpairsTypx{} \textsc{typx}, and
\RCpairsRw{} \textsc{rewiring}); the supplement lists each pair.
The complete exploratory list was not specified in the six-hypothesis
pre-registration; every metric enters at least once.
The six pre-registered hypotheses form a separate family
(Section~\ref{sec:holdout}).

\paragraph{Data construction and reproducibility.}
\label{sec:data}
Annotations are matched to the measured assistant response by identifier, and
each test uses the turn pairs for which both its metric and its label are
available.
Several metrics have no defined value at the start of a conversation
(differences, windows, and quantities built on them).  The main analysis
treats these initialization values as missing, instead of reading them as
measurements, and rebuilds the dependent quantities from their measured inputs.
This construction was chosen after initial analyses and is not
pre-specified; keeping the initialization values as computed gives a failure
share of \RRPrefixMetricOnlyFailure\% instead of \RNirBHJoint\% and confirms
more of the pre-registered hypotheses (Appendix~\ref{app:results}, which also
reports two intermediate constructions).
The per-pair inventory (one row per test with $r$, pooled $p$, both
autocorrelations, effective sample sizes, $p_{\mathrm{boot}}$, and
$p_{\mathrm{final}}$) ships in the supplement together with validation
scripts for the reported tables.  The conversations themselves are not
released, so the supplement does not recompute the metric values from raw
text.

\subsection{Results}
\label{sec:results}

Table~\ref{tab:main} presents the central finding for the \textsc{malicious}
label: all BH-screened pairs grouped by metric category and evaluated under
the dependence-aware criterion.

\begin{table}[t]
\centering
\footnotesize
\begin{tabular}{@{}lrrrr@{}}
\toprule
\textbf{Category} & \textbf{\#Sig\textsubscript{BH}} & \textbf{\#Rob.} & \textbf{FS\,(\%)} & $\boldsymbol{\bar{\rho}_x}$ \\
\midrule
Cumulative accum. & 2 & 1 & 50.0 & 0.91 \\
Normalized cumul. & 1 & 0 & 100.0 & 0.76 \\
Rolling aggr. (mixed W) & 1 & 1 & 0.0 & 0.67 \\
Weighted variants & 2 & 1 & 50.0 & 0.61 \\
Thermodynamic & 6 & 2 & 66.7 & 0.37 \\
Lexical/structural & 4 & 2 & 50.0 & 0.31 \\
Interaction & 4 & 0 & 100.0 & 0.26 \\
Frame distance & 0 & 0 & --- & 0.19 \\
Embedding velocity & 6 & 3 & 50.0 & 0.18 \\
Impulse/indicator & 4 & 4 & 0.0 & 0.12 \\
Compression (NCD) & 1 & 0 & 100.0 & 0.11 \\
Timestamp/recovery & 2 & 2 & 0.0 & 0.05 \\
Directional consist. & 1 & 1 & 0.0 & $-$0.01 \\
\midrule
\textbf{Total (\textsc{malicious})} & \textbf{34} & \textbf{17} & \textbf{50.0} & \\
\textbf{All three labels} & \textbf{\RNbh} & \textbf{\GBootScreenH} & \textbf{\GBootFSH} & \\
\bottomrule
\end{tabular}
\caption{%
  Dependence-aware survival by metric category for the \textsc{malicious}
  label (\RNnConv{} conversations, \RNnPairs{} turn pairs).
  \textbf{\#Sig\textsubscript{BH}}: passes pooled BH screening at
  $q{<}0.05$ (\RNbhMal{} \textsc{malicious} tests of \RNbh{} total).
  \textbf{\#Rob.}: nominally confirmed, $p_{\mathrm{final}} < 0.05$.
  \textbf{FS}: failure share (Eq.~\ref{eq:inflation-rate}).
  $\bar{\rho}_x$: mean lag-1 metric autocorrelation over the label's tests
  in the category (it can therefore differ from the family mean over all
  metrics in Table~\ref{tab:autocorrelation}).
  Bottom rows: totals for \textsc{malicious} and across all three labels
  (per-label robust counts \RNlabMal/\RNlabTypx/\RNlabRw{} for
  \textsc{malicious}/\textsc{typx}/\textsc{rewiring}).%
}
\label{tab:main}
\end{table}

Several patterns emerge.
First, the failure share varies across categories from 0\% to 100\%, but the
ordering of the categories is not reproduced when individual tests are split
at the same autocorrelation threshold (Section~\ref{sec:inflation-rate}).
Second, the two components of the criterion play unequal roles on this
corpus.  The joint analytic correction alone would pass all \RNbh{} screened
pairs; the conversation-level bootstrap passes \RNrobJoint{} at the nominal
threshold, and these are the nominally confirmed pairs under the joint form,
the conservative bound, and the bootstrap alone.
The lag-1 formula captures persistence from turn to turn; the bootstrap
additionally captures the differences between conversations, which carry a
large part of the variation of the cumulative metrics and of the label itself
(Section~\ref{sec:why-vulnerable}).
Which dependence mechanism produces the failures is not isolated by this
comparison.
The analytic component binds elsewhere: it withholds one association that the
pooled test does not support but the bootstrap alone would pass, and in the
smaller hold-out tiers its conservative form removes 12 and 14
bootstrap-significant tests (Section~\ref{sec:holdout}).
Third, robustness belongs to the metric--label pair, not to the metric: the
compression distance fails for the \textsc{malicious} label
($p_{\mathrm{final}} = 0.128$) and is robust for \textsc{typx} ($r = 0.131$,
one of twenty tests without a single opposite-sign replicate among the
2{,}000 resamples; its sign changes in at least one leave-one-user-out
analysis, Table~\ref{tab:full}).
Fourth, two metrics of the battery select their fitting rows with the help of
the labels (Appendix~\ref{app:metrics}), so their four tests are not
independent evidence; excluding them and reapplying both BH steps leaves 34
of 63 screened pairs (failure share 46.0\%) and the same \RNbhPfinalJoint{}
discoveries.
A further redundancy concerns \texttt{Psi}, an affine transform of
\texttt{Entropy}: omitting its test and reapplying both BH steps leaves
\RCnoPsiConf{} of \RCnoPsiScr{} screened pairs (failure share
\RCnoPsiFail\%) and the same \RCnoPsiBH{} discoveries.
Excluding all five tests and reapplying both BH steps leaves
\RCnoBothConf{} of \RCnoBothScr{} screened pairs (failure share
\RCnoBothFail\%) and again the same \RCnoBothBH{} discoveries; the main
results keep all \RNnTests{} tests.
Each test draws its own 2{,}000 resamples, so the tests on \texttt{Entropy}
and \texttt{Psi}, identical up to sign, obtain bootstrap values of
\RCpbootEntropy{} and \RCpbootPsi; near
0.05 the Monte Carlo standard error of $p_{\mathrm{boot}}$ is about \RCmcSE,
and two screened pairs lie within two standard errors of the threshold
(\RCmcNearLow{} and \RCmcNearHigh, both counted as failures).
The same Monte Carlo error applies at the BH boundary: the combined values
at ranks \RCbhLastRank{} and \RCbhNextRank{} (\RCbhLastP{} and \RCbhNextP)
lie within one standard error (about \RCbhSE) of their thresholds
(\RCbhLastThr{} and \RCbhNextThr); the set of BH discoveries could
therefore change with different resamples.

\begin{figure}[t]
\centering
\includegraphics[width=0.62\linewidth]{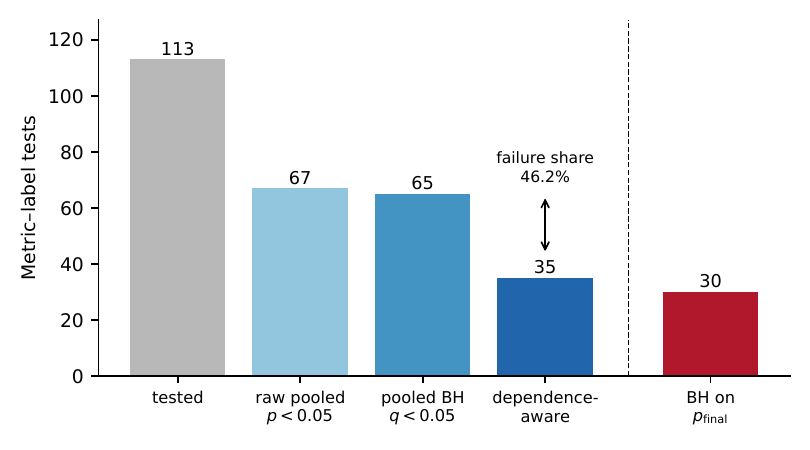}
\caption{%
  Correction funnel on the full corpus.  The failure share
  (Eq.~\ref{eq:inflation-rate}) is the gap between the BH-screened and the
  dependence-aware set.  The separate bar shows the discoveries of the primary
  multiple-testing procedure, BH applied to $p_{\mathrm{final}}$.%
}
\label{fig:funnel}
\end{figure}

\paragraph{Post hoc check: resampling budget.}
The results above use $B = 2{,}000$ resamples per test.
In a check planned after these results were known, each test was extended
to \RBtotalB{} resamples (the stored 2{,}000 plus \RBaddB{} additional
resamples of whole conversations from a separate random stream per test),
and $p_{\mathrm{boot}}$ and $p_{\mathrm{final}}$ were recomputed from all
of them.
The two budgets share the data and the first 2{,}000 resamples, so their
results are not independent; the pooled screen does not depend on the
budget.
The same \RBconfNew{} of \RNbh{} screened pairs pass at the nominal
threshold (failure share \RBfailNew\%).
BH on the combined values yields \RBbhNew{} discoveries instead of
\RNbhPfinalJoint: all \RNbhPfinalJoint{} remain and
\textsc{malicious}--\texttt{MAS} joins (\RBgainOldB{} of 2{,}000 and
\RBgainNewB{} of \RBtotalB{} resamples of opposite sign;
$p_{\mathrm{final}} = \RBgainOldP$ and $\RBgainNewP$).
Each of the three reduced batteries above again yields the discoveries of
the full battery.
The lead example of Section~\ref{sec:intro} still fails
($p_{\mathrm{final}} = \RBleadOldP$ and $\RBleadNewP$), the tests on
\texttt{Entropy} and \texttt{Psi} obtain bootstrap values of \RBentNewP{}
and \RBpsiNewP, and no screened pair lies within two Monte Carlo standard
errors of 0.05.
With \RBtotalB{} resamples the smallest attainable bootstrap value is
\RBminP, below the first BH threshold; at the new boundary the value at
rank \RBedgeRank{} (\RBedgeP) lies within one standard error (about
\RBedgeSE) of its threshold (\RBedgeThr), so the decision at that rank
could again change with different resamples.
No other analysis was rerun: Tables~\ref{tab:main}, \ref{tab:full}
and~\ref{tab:variants}, Figure~\ref{fig:funnel},
Sections~\ref{sec:users} and~\ref{sec:holdout} and the other post hoc
checks refer to the decisions obtained with the original budgets (2{,}000
resamples per test unless stated), and the simulated error rates and power
(Appendix~\ref{app:simulation}) refer to $B = 2{,}000$.

Figure~\ref{fig:funnel} visualizes the correction funnel.
An alignment check counts, over the \RCrotCells{} of \RCrotPlanned{}
conversation--test combinations with at least ten available turn pairs and
two non-constant series, the
within-conversation correlations that are nominally significant, and compares
the count with a circular-rotation null that preserves the cyclic order of both
series while shifting the labels relative to the metric: \RNobsSig{} observed against
$\RNrotMean \pm \RNrotSD$ under rotation ($\RNrotFac\times$; no exceedance in
500 rotations).
A within-conversation label shuffle, which destroys the label series'
dependence, gives $\RNshufMean \pm \RNshufSD$ and would overstate the same
excess as $\RNshufFac\times$.

\paragraph{Post hoc checks: turn position and conversation level.}
Two descriptive checks were computed after the main analysis from the stored
turn-level values, without resampling; they change no decision reported
above, and the supplement lists the correlations for all \RNnTests{} tests.
The first concerns turn position: in the pooled data the \textsc{malicious}
label is more frequent at later turn positions, and cumulative and windowed
metrics change with position by construction.
Of the \RNrobJoint{} nominally confirmed pairs, \RCposLateKept{} retain at
least two thirds of their pooled correlation when only turn pairs from
position 20 onward (counted from zero) are used, and \RCposBinKept{} do so
after the means of nine position bins are removed from both series.
Only the \textsc{malicious}--\texttt{delta\_NC2} correlation changed sign in
both position checks.
For this pair, one of the \RNbhPfinalJoint{} BH discoveries, the correlation
was $\RCwindowR$ over all available turn pairs and $\RCwindowLateR$ over the
\RCwindowLateN{} pairs for which both windows were full; the BH decision was
not recomputed for this subset.
Resampling entire conversations preserves this position-related pattern.
\texttt{NC2} is computed over a trailing window that fills during the first
twenty turns; while it fills, \RCwindowOpenPos{} of \RCwindowOpenN{}
differences are positive, and in the pooled data the label is less frequent
at these positions than later (\RCwindowOpenLabel\% against
\RCwindowFullLabel\%).
The second check centers both series within each conversation.
\RCwithinKept{} of the \RNrobJoint{} pairs then retain at least two thirds
of their pooled correlation, \RCwithinHalf{} between one half and two
thirds, and \RCwithinBelow{} less than half; \RCwithinFlip{} change sign.
Most pooled associations that pass the dependence-aware procedure are
therefore markedly weaker within conversations than in the pooled data;
whether the within-conversation correlations differ from zero was not
tested.

\subsection{User Sensitivity}
\label{sec:users}

All \RNnConv{} conversations come from four users, and conversation-level
resampling treats conversations as the independent unit; with a dominant
user this assumption deserves its own audit.
Three analyses quantify it; they were run with the conservative-bound form of
the criterion on the raw pooled base (baseline failure share \RNirRawMO\%).
First, leave-one-user-out: dropping any single user gives a failure share
between \RNlouoLo\% and \RNlouoHi\%.
Without the dominant user, \RClouoRobust{} of 40 raw pooled-significant pairs
remain robust (failure share \RNlouoHi\%).
Of those \RClouoRobust, \RClouoOverlap{} are among the \RNrobMO{} robust pairs in the full
corpus under the same criterion.
Second, under a user-level cluster bootstrap \RNclusterRob{} of the
\RNpooledRaw{} pairs are robust; with four clusters this is a coarse
sensitivity check, not a test for a wider population of users.
Third, the qualitative finding holds under every exclusion (at least
\RNlouoLo\% of the pooled-significant pairs fail), while individual pairs are
not robust to the choice of users:
eight of the \RNrobMO{} robust pairs change sign in at least one
leave-one-user-out run (marked in Table~\ref{tab:full}), and 15 of them are
not robust under the user-level bootstrap, which in turn passes six other
pairs.
User-robustness is therefore reported per pair in the released inventory
rather than claimed globally.
The public corpora (Section~\ref{sec:publicdata}), with different speakers,
languages, and constructs, address the same concern from outside this user
pool.

\subsection{Hold-Out Replication on the Pre-Registered Split}
\label{sec:holdout}

A reanalysis of the pre-registered hold-out split compares how often
dependence-aware and pooled-only findings replicate.  Tier~A (\RNtierAconv{}
strictly unseen conversations, \RNtierApairs{} turn pairs) is primary; Tier~B
(\RNtierBconv{} conversations, \RNtierBpairs{} pairs) contains Tier~A and 14
pilot-touched conversations.
In the \RNexplConv-conversation exploratory set, 56 tests are raw-pooled
significant: 35 dependence-aware and 21 pooled-only (sets specific to this
split).
Groups and the strict replication criterion use the conservative-bound form
of $p_{\mathrm{final}}$.

\begin{table}[t]
\centering
\small
\begin{tabular}{@{}llcccc@{}}
\toprule
 & & \multicolumn{2}{c}{\textbf{Tier A} (\RNtierAconv{} conv.)} & \multicolumn{2}{c}{\textbf{Tier B} (\RNtierBconv{} conv.)} \\
\textbf{Criterion in hold-out} & \textbf{Exploratory status} & repl. & rate & repl. & rate \\
\midrule
Strict ($p_{\mathrm{final}}<.05$) & dependence-aware ($n=35$) & 10 & 28.6\% & 13 & 37.1\% \\
 & pooled-only ($n=21$) & 2 & 9.5\% & 1 & 4.8\% \\
\midrule
Bootstrap leg only & dependence-aware & 16 & 45.7\% & 21 & 60.0\% \\
 & pooled-only & 4 & 19.0\% & 3 & 14.3\% \\
\bottomrule
\end{tabular}
\caption{%
  Hold-out replication by exploratory status, per tier and criterion.
  A finding replicates if it has the same direction and is significant in the
  hold-out under the stated criterion (``strict'': $p_{\mathrm{final}}$ in its
  conservative-bound form; ``bootstrap only'': $p_{\mathrm{boot}}$).
  Tier~B contains Tier~A.%
}
\label{tab:holdout}
\end{table}

Four observations stand out.
\textbf{(i)}~Dependence-aware findings replicate more often than pooled-only
findings in both tiers and under both criteria (Table~\ref{tab:holdout}):
under the strict criterion, \RNrepAstr\% versus \RNrepAstrP\% in Tier~A and
\RNrepBstr\% versus \RNrepBstrP\% in Tier~B.
The 56 tests share conversations, labels, and related metrics, and the
pooled-only group is small; we therefore report the rates descriptively and
attach no significance statement to their ratio.
In Tier~A, \RChoSignRobA{} of \RChoRobN{} and \RChoSignPoolA{} of
\RChoPoolN{} pairs, respectively, retained their exploratory sign
(\RChoSignRobB{} and \RChoSignPoolB{} in Tier~B); \RChoStrictRobA{} and
\RChoStrictPoolA{} met the strict replication criterion in Tier~A.
This count without a significance threshold was made post hoc; a pair that
does not meet the criterion again is not thereby shown to be absent or
reversed.
The largest user accounted for \RCtierALargestPairs{} of the
\RNtierApairs{} turn pairs in Tier~A (\RCtierALargest\%).
\textbf{(ii)}~A logistic model of hold-out replication on correction status
\emph{and} exploratory $|r|$ gives a positive coefficient for the robust
indicator in both tiers and a negative coefficient for $|r|$.  Its
ordinary-model $z$-statistics ($\RNzA$ in Tier~A, $\RNzB$ in Tier~B) ignore
the dependence between tests and are not calibrated evidence for a
replication advantage.  Within each effect-size tercile, robust pairs
replicate at least as often.
\textbf{(iii)}~The six pre-registered directional hypotheses, tested as
specified (Spearman correlation, one-sided conversation-level bootstrap, Holm
at $\alpha = 0.05$), yield \RNholmAConfirmed/6 confirmations in Tier~A and
\RNholmBConfirmed/6 in Tier~B; all twelve estimated directions match the
predicted ones.
With the meta-level criterion below also met, this is the outcome the
pre-registration defines as partial confirmation (two or three of six); full
confirmation would have required four.
\textbf{(iv)}~The pre-registered meta-level criterion requires more than
$2\,(0.05\times\RNnTests)=11.3$ bootstrap-significant associations, twice the
number expected under the null; the hold-out yields 30 in Tier~A and 33 in
Tier~B (\RNmetaA{} and \RNmetaB{} under the strict criterion).
This reports the pre-registered count criterion, not a calibrated test of excess discoveries.

\subsection{The EWMA Trade-off}
\label{sec:ewma}

One result illustrates the central tension of this paper.
Exponential smoothing (EWMA, $\alpha = 0.3$) applied to semantic velocity
raises its point-biserial correlation with the \textsc{malicious} label from
$r = \RNrawR$ to $r = \RNewmaR$.
Simultaneously, it raises lag-1 autocorrelation from $\rho = \RNrawRho$
to $\rho = \RNewmaRho$.  The gain in the correlation and the change in
uncertainty have to be assessed separately.

\paragraph{A numerical resolution.}
Table~\ref{tab:ewma} lists the interval of the smoothed metric under each
inference variant.  The conversation-level bootstrap interval is more than
twice as wide as the naive one; relative to the joint approximation the naive
interval is \RNciShrink\% narrower, relative to the conservative bound
\RNciShrinkMO\%.  All intervals exclude zero: the association is supported
under every variant.  What the naive analysis overstates is the
\emph{precision} of the evidence.

\begin{table}[t]
\centering
\small
\begin{tabular}{@{}lcccc@{}}
\toprule
  & $r$ & $\bar{\rho}_1$ & $n_{\mathrm{eff}}$ & 95\% CI \\
\midrule
Raw metric, conversation bootstrap  & \RNrawR  & \RNrawRho  & --- & $[0.018,\; 0.078]$ \\
EWMA, conversation bootstrap        & \RNewmaR & \RNewmaRho & --- & $[0.047,\; 0.137]$ \\
EWMA, joint approximation           & \RNewmaR & \RNewmaRho & \RNewmaNeff & $[\RNewmaCIlo,\; \RNewmaCIhi]$ \\
EWMA, conservative bound            & \RNewmaR & \RNewmaRho & \RNewmaNeffMO & $[\RNewmaMOlo,\; \RNewmaMOhi]$ \\
\textit{EWMA, naive}                & \textit{\RNewmaR} & \textit{---} & \textit{\RNewmaN} & \textit{$[\RNewmaNaivLo,\; \RNewmaNaivHi]$} \\
\bottomrule
\end{tabular}
\caption{The EWMA trade-off (semantic velocity, \textsc{malicious} label).
Bootstrap rows: 95\% percentile intervals from 2{,}000 whole-conversation
resamples; remaining rows: Fisher-$z$ intervals at the stated
$n_{\mathrm{eff}}$.  Smoothing raises the observed correlation
($r$: $\RNrawR \to \RNewmaR$) and the autocorrelation
($\rho$: $\RNrawRho \to \RNewmaRho$); the naive interval (italicized) is the
narrowest.}
\label{tab:ewma}
\end{table}

\section{Evidence from Public Corpora}
\label{sec:publicdata}

The scope limits of the main study---four users, one language, private
data, and a bespoke metric suite---are best addressed with public data.  We
therefore ran the same procedure (pooled point-biserial tests, BH screening,
joint dependence correction, conversation-level block bootstrap) with
\emph{standard} conversational metrics on three public corpora with
per-utterance human labels, and we add a fourth human-annotated corpus on
which the effect of the resampling unit can be shown directly.
All four corpora contain conversations between humans: they test the
statistical argument, not the manipulation construct of the main study.

\subsection{Three Corpora with Per-Utterance Labels}
\label{sec:public-three}

\begin{itemize}[nosep]
\item \textbf{DailyDialog} \citep{li2017dailydialog} (a random sample of 2{,}000 of the dialogues with at least eight utterances,
  22{,}244 utterances; emotion
  labels, tested as non-neutral vs.\ neutral, and dialogue-act labels, tested
  as question vs.\ rest);
\item \textbf{MELD} \citep{poria2019meld} (the 1{,}015 dialogues with at least six utterances, 12{,}443 utterances;
  sentiment labels,
  tested as negative vs.\ rest);
\item \textbf{Switchboard} \citep{stolcke2000dialogue} (the 1{,}150 telephone conversations with at least 20 utterances,
  122{,}563 utterances; dialogue-act labels, tested as question vs.\ rest
  and backchannel vs.\ rest).
\end{itemize}

The battery contains 12 standard measures: turn length, type-token ratio,
lexicon sentiment (a per-turn classifier score), unigram surprisal, embedding
velocity, rolling sentiment ($W{=}5$ and $W{=}20$), cumulative sentiment (sum
and mean), EWMA velocity, and first differences of sentiment and turn length,
so that the main construction classes (per-turn, rolling, cumulative,
smoothed, differenced) are represented by community-standard quantities and
not by our suite.
These measures follow the original utterance order, including speaker
changes, and rolling means use partial initial windows: for a dialogue of at
most 20 utterances the $W{=}20$ mean equals the cumulative mean, so the two
nearly coincide on DailyDialog and MELD (mean lengths of 11 and 12
utterances).
The two labels on DailyDialog and Switchboard yield 24 tests each; MELD
yields 12.

\paragraph{Design-class ordering replicates.}
Table~\ref{tab:community} reports the measured mean lag-1 autocorrelation
per design class on all three corpora.  The same coarse ordering by
construction appears on every public corpus: cumulative sums are the most persistent
(0.58--0.61 on the short-dialogue corpora, 0.96 on Switchboard, whose
conversations are an order of magnitude longer), rolling windows, EWMA, and
cumulative means follow (0.39--0.83), per-turn quantities are weakly autocorrelated (between
$-0.22$ and $+0.15$), and differenced metrics are anti-persistent ($-0.44$ to
$-0.58$).
The persistence of a metric is thus largely a property of how it is
constructed.

\begin{table}[t]
\centering
\footnotesize
\begin{tabular}{@{}lccc@{}}
\toprule
Design class & DailyDialog & MELD & Switchboard \\
\midrule
Cumulative (sum) & 0.61 & 0.58 & 0.96 \\
Rolling window (W=20) & 0.39 & 0.40 & 0.83 \\
Rolling window (W=5) & 0.47 & 0.48 & 0.78 \\
EWMA & 0.44 & 0.44 & 0.74 \\
Cumulative (mean) & 0.39 & 0.40 & 0.72 \\
Embedding velocity & $-$0.06 & $-$0.06 & 0.15 \\
Per-turn classifier & $-$0.05 & $-$0.06 & 0.01 \\
Lexical & $-$0.12 & $-$0.08 & $-$0.14 \\
Per-turn scalar & $-$0.14 & $-$0.07 & $-$0.22 \\
Differential & $-$0.48 & $-$0.44 & $-$0.58 \\
\bottomrule
\end{tabular}
\caption{Measured mean lag-1 autocorrelation of standard metric design
classes on three public corpora.  Per-metric values and the full test
inventories are in the supplement.}
\label{tab:community}
\end{table}

\paragraph{Failure shares.}
Unlike ours, the label series of these corpora carry no positive
persistence: their mean per-conversation lag-1 autocorrelation is near zero
for DailyDialog emotion ($+0.03$), MELD sentiment ($0.00$), and Switchboard
questions ($-0.04$), and negative for DailyDialog questions and Switchboard
backchannels (about $-0.35$ and $-0.31$).
On these corpora the two procedures largely agree on rejection decisions: the
failure share is \GPublicDailyFS\% on DailyDialog (23/23 BH-screened pairs
remain), \GPublicMeldFS\% on MELD (11/11), and \GPublicSwitchFS\% on
Switchboard (18/19).
With the bootstrap retained, the conservative bound gives total failure
shares of 1/11 on MELD (\RNirMeld\%) and 3/19 on Switchboard (\RNirSwbd\%).
Dependence between turns is therefore not sufficient by itself to change
pooled decisions: with a thousand or more conversations and labels with
near-zero or negative mean lag-1 autocorrelation, significance decisions
largely coincide, which does not imply equal uncertainty or calibrated
pooled tests.
The corpora differ from ours in labels, lengths, metrics, and the unit of
observation (single utterances instead of prompt--response pairs), so the contrast
with \RNirBHJoint\% compares settings and is not a controlled experiment; the
controlled counterpart is the simulation, which varies the serial dependence
of both series (Appendix~\ref{app:simulation}).
In Switchboard, 303 of the 439 speakers appear in more than one conversation,
so its conversations are not independent samples of speakers.
Two further candidate corpora were excluded for material reasons:
ProsocialDialog carries only 2--3 labeled turns per dialogue, and OASST
conversation paths are too short for the rolling variants.

\subsection{The Resampling Unit on Human-Annotated Counselling Dialogues}
\label{sec:annomi-public}

AnnoMI~\citep{wu2023annomi} contains 133 transcripts from 119 source videos
of motivational-interviewing demonstrations (9{,}699 utterances) with expert
annotations.  We analyze client and therapist utterance sequences separately,
with three label contrasts (client change talk, client sustain talk,
therapist reflection) and eight simple measures of utterance length and
lexical variety (logarithmic token count, type--token ratio, rolling means
over 5 and 20 utterances, EWMA, cumulative mean, cumulative sum, and first
difference; Appendix~\ref{app:public-extension}).
In contrast to the three corpora above, two of the three label series are
positively persistent (mean lag-1 autocorrelation $+0.206$ for change talk,
$+0.141$ for sustain talk, $+0.034$ for reflection; estimator of
Appendix~\ref{app:public-extension}, which differs from that of
Table~\ref{tab:community} and the main study, Appendix~\ref{app:metrics}).
Because AnnoMI is analyzed in separate role sequences and with complete
windows, the autocorrelations of the two subsections are not a controlled
comparison of persistence.

For all 24 metric--label pairs we compare two 95\% percentile intervals for
the same pooled correlation, computed on the same observations: one from
resampling individual utterances, one from resampling whole source videos
(19{,}999 resamples each).  Only the resampling unit differs.
The video-level interval is wider by a median factor of \PEVideoIIDMedian{}
(range \PEVideoIIDMin--\PEVideoIIDMax; wider in 23 of the 24 pairs); zero is
excluded by 22 of the 24
utterance-level intervals and by 17 of the video-level intervals.
Applying the diagnostic of Section~\ref{sec:inflation-rate} to the same pairs
(pooled test with BH screening, then the video-level interval), 4 of 21
screened pairs fail: a failure share of 19.0\%.
Sustain talk paired with the 20-utterance rolling mean loses its
utterance-level zero exclusion but lies outside the pooled BH screen, which
accounts for the difference between five interval losses and four screened
failures.

\begin{table}[t]
\centering\small
\begin{tabular}{@{}lrr@{}}
\toprule
Construction & Pairs & Video/utterance width ratio \\
\midrule
First difference & 3 & 0.989--1.347 \\
Log length and type--token ratio & 6 & 1.323--1.862 \\
Rolling means, EWMA, cumulative mean & 12 & 1.415--2.680 \\
Cumulative sum & 3 & 1.776--4.005 \\
\bottomrule
\end{tabular}
\caption{AnnoMI: ratio of the width of the video-level interval to the width
of the utterance-level interval, by construction of the measure (all 24
pairs; Appendix~\ref{app:public-extension} lists each pair).}
\label{tab:public-construction}
\end{table}

Width ratios differ descriptively across constructions, with overlapping
ranges and feature-specific observation sets
(Table~\ref{tab:public-construction}).
Each video/utterance comparison uses the same observations; comparisons across
features do not isolate construction from sample selection.
First differences stay between 0.99 and 1.35.
Untransformed per-utterance measures lie between 1.32 and 1.86 although their
lag-1 autocorrelation is near zero.
This is consistent with a contribution from variation shared within a recording
(Section~\ref{sec:why-vulnerable}); the comparison does not separate that contribution from
persistence within recordings.
Cumulative sums show the largest ratios, up to \PEVideoIIDMax, and all five pairs that lose their zero exclusion involve
rolling means or cumulative sums.

This analysis is not pre-registered.  The
24 pairs were fixed before any interval was computed; the utterance-level
bootstrap was added as a like-for-like reference after a first comparison
with Fisher intervals, which gives a similar median ratio
(\PEVideoFisherMedian).
The recordings are demonstrations, not a clinical sample, and the same
persons may appear in more than one video.

\section{Design Principles and Recommendations}
\label{sec:recommendations}

The empirical pattern from Sections~\ref{sec:empirical}
and~\ref{sec:publicdata} motivates five design principles for turn-level
conversational analysis.

\begin{enumerate}[leftmargin=*,itemsep=2pt]

\item \textbf{Prefer differences over levels when the question is
  turn-level.}
  Compute $\Delta x_t = x_t - x_{t-1}$ instead of raw $x_t$ when change is
  the target.  First-differencing a per-turn quantity can yield weakly
  persistent or anti-persistent increments (e.g.\ $\bar\rho_1 =
  \RNmetRhoMin$ for the differenced compression metric); on AnnoMI, first
  differences are the construction least affected by the resampling unit
  (Table~\ref{tab:public-construction}).
  First differences of windowed quantities are not covered by this rule:
  \texttt{delta\_NC2} has $\bar\rho_1 = \RCwindowRho$, its differences are
  almost all positive while its window fills, and its pooled association
  with the label depends on turn position (Section~\ref{sec:results}).
  A question about accumulated state still calls for a cumulative measure.

\item \textbf{Report lag-1 autocorrelation of \emph{both} series alongside
  every turn-level result.}
  The product $\bar\rho_x\bar\rho_y$ enters the AR(1) approximation of the
  effective sample size (Eq.~\ref{eq:chelton-joint}), and label dependence is easily overlooked.
  Report as well how much of each variable's variation lies between
  conversations: low autocorrelation within conversations can coexist with
  strong clustering.

\item \textbf{Account for the sampling unit.}
  Use inference that respects the sampling units and the dependence within
  conversations.  Conversation-level block bootstrap, cluster-robust
  standard errors \citep{cameron2015practitioner}, or appropriately specified
  multilevel models are possible approaches; state their sampling
  assumptions.
  Where conversations share users, speakers, or recordings, the higher unit
  should be examined as well (Sections~\ref{sec:users}
  and~\ref{sec:annomi-public}).

\item \textbf{Run multiple-testing control on dependence-aware
  $p$-values.}
  Apply BH to $p_{\mathrm{final}} = \max(p_{\mathrm{Chelton}},
  p_{\mathrm{boot}})$, not to pooled $p$-values
  (Section~\ref{sec:protocol}).  If a pooled screen is used for
  computational reasons, label it as a screen; pooled tests on their own
  merit only an \emph{exploratory} label.

\item \textbf{Report the failure share as a transparency metric.}
  $\IR = 1 - n_{\text{robust}} / n_{\text{pooled-sig}}$, with $n_{\text{robust}}$
  counted inside the pooled screen (Eq.~\ref{eq:inflation-rate}),
  communicates in a single number how much of an analysis's pooled
  significance fails under dependence-aware inference.  In our study,
  $\IR = \RNirBHJoint\%$; Section~\ref{sec:publicdata} reports the shares
  measured on public corpora.
  The share measures disagreement between two procedures, not the fraction
  of false findings.

\end{enumerate}

\medskip
These principles are summarized in a concrete checklist that we
recommend authors and reviewers adopt for any study reporting turn-level
correlations or regressions in multi-turn conversational data.

\begin{figure}[!ht]
\begin{tcolorbox}[colback=gray!5, colframe=black!70, title={\textbf{Checklist: Before You Publish Turn-Level Results}}, fonttitle=\small\bfseries, boxrule=0.5pt, arc=2pt, left=4pt, right=4pt, top=3pt, bottom=3pt]
\small
\begin{enumerate}[leftmargin=*,itemsep=1pt,topsep=2pt]
\item[$\square$] \textbf{Sampling unit.} State which unit is treated as independent (conversation, user, recording) and what the reported association is meant to estimate.
\item[$\square$] \textbf{Autocorrelation audit.} Compute lag-1 $\rho$ for every turn-level metric \emph{and} for every label series; pooled $p$-values are least trustworthy where both are persistent. Report the share of each variable's variation that lies between conversations.
\item[$\square$] \textbf{Effective sample size.} Report $n_{\text{eff}} = n (1{-}\bar{\rho}_x\bar{\rho}_y)/(1{+}\bar{\rho}_x\bar{\rho}_y)$ alongside nominal $n$, as an approximation of the serial component.
\item[$\square$] \textbf{Cluster-aware inference.} Use conversation-level block bootstrap, cluster-robust SEs, or mixed-effects models. Label pooled tests as \emph{exploratory}.
\item[$\square$] \textbf{Multiple testing.} Apply BH to dependence-aware $p$-values; choose the resampling count so that the smallest attainable $p$-value lies below the first BH threshold $q/m$ (with $B=2{,}000$ and $m=\RNnTests$ here it lies between the second and the third, with the \RBtotalB{} resamples of Section~\ref{sec:results} below the first; Section~\ref{sec:protocol}).
\item[$\square$] \textbf{Failure share.} Report $\IR = 1 - n_{\text{robust}}/n_{\text{pooled-sig}}$, with $n_{\text{robust}}$ counted inside the pooled screen, as a transparency metric.
\item[$\square$] \textbf{Metric design check.} For cumulative and rolling metrics, verify with conversation-level aggregation; consider first-differencing ($\Delta x_t$) when change is the target.
\end{enumerate}
\end{tcolorbox}
\end{figure}

\paragraph{When cumulative metrics are appropriate.}
Cumulative metrics are not inherently flawed: they are
the natural unit of analysis for \emph{session-level} questions (``Is
this conversation, as a whole, manipulative?'').  The problem arises only
when cumulative scores are tested against turn-level labels with pooled
inference, because thousands of measurements along a limited number of
conversation histories are then counted as thousands of independent
observations.  Even at the turn level, cumulative and rolling metrics can
survive the dependence-aware criterion when the association holds across
conversations---one of the two screened cumulative-accumulator pairs and the
screened rolling aggregate do so in this corpus (Table~\ref{tab:main}).
The principle: \textbf{match the grain
of the metric to the grain of the label, and match the inference to the
dependence of both series.}

\section{Conclusion}
\label{sec:conclusion}

Serial dependence in multi-turn conversational data is common and can be
consequential, and pooled turn-level tests that ignore it are documented in
the evaluation literature we audited.

On our corpus, \RNirBHJoint\% of the associations that pass pooled BH
screening fail to remain significant under a dependence-aware procedure
that accounts for the autocorrelation of both the metric and the label
series and resamples whole conversations; the conversation-level bootstrap
determines these decisions and preserves the differences between
conversations as well as their internal histories.
On three public corpora whose labels have near-zero or negative mean lag-1
autocorrelation the same procedure
yields \GPublicDailyFS\%, \GPublicMeldFS\%, and \GPublicSwitchFS\%, and on
human-annotated counselling dialogues, changing only the resampling unit
widens the intervals of 24 correlations by a median factor of
\PEVideoIIDMedian.
In simulation, pooled inference stays near its nominal level when either
series is serially independent and exceeds it increasingly as both become
dependent.
The correction itself is simple---a conversation-level block bootstrap
plus the two-series effective-degrees-of-freedom formula, with BH applied
to the corrected $p$-values---and adds little code to a standard
correlation analysis; we release the inference code, the complete per-pair
inventory with validation scripts, and definitions of all \RNnMetrics{}
metrics.
In the reanalysis of the pre-registered hold-out split, findings classified
as dependence-aware under the conservative-bound form of the criterion
replicate more often than pooled-only findings in both tiers (\RNrepAstr\%
versus \RNrepAstrP\% in the primary tier), and two of the six pre-registered
directional hypotheses are confirmed in each tier.

\paragraph{Future work.}
Three directions are immediate: public human--LLM conversation data with
more users; integration into dialogue assessment toolkits
\citep{mehri2020usr,deriu2021survey}; and analytic cluster-robust variance
estimators for streaming use, where bootstrap resampling is costly.

\section*{Ethics Statement}

This study analyzes conversations between human users and four LLM platforms (ChatGPT, DeepSeek, Gemini, and Claude).
All participants gave informed consent for their conversation data to be used in this research.
Conversations are not released to protect participant privacy.
The LLM-as-judge annotation protocol was designed to evaluate conversational dynamics, not to profile or classify individual users.
The statistical correction methodology we propose is domain-agnostic and does not raise dual-use concerns.

\section*{Limitations}

\paragraph{Sample scope.}
Our empirical demonstration draws on four users, one language (German), and four LLM platforms (ChatGPT, DeepSeek, Gemini, Claude); the largest user contributes \RNpersonAShare\% of turn pairs (Table~\ref{tab:users}).
The specific failure share is not a claimed constant: on three public corpora the same procedure yields \GPublicDailyFS\%, \GPublicMeldFS\%, and \GPublicSwitchFS\% (Section~\ref{sec:publicdata}), and within our corpus leaving out any single user gives between \RNlouoLo\% and \RNlouoHi\% under the conservative-bound form on the raw pooled base, where the full corpus gives \RNirRawMO\% (Section~\ref{sec:users}).
Individual associations are not robust to the choice of users, and four users cannot support inference about a population of users.
The main data construction was chosen after initial analyses; the alternative gives \RRPrefixMetricOnlyFailure\% (Section~\ref{sec:setup}; Appendix~\ref{app:results}).

\paragraph{Annotation.}
Turn-level labels were produced by an LLM-as-judge pipeline (Appendix~\ref{app:annotation}).
An expert review by a computational linguist confirmed annotation quality on a stratified subsample, but a formal inter-rater reliability study was not conducted.
The label series are themselves autocorrelated (median per-conversation lag-1 of \RNrhoYmal{} for \textsc{malicious}, \RNrhoYtypx{} for \textsc{typx}, \RNrhoYrw{} for \textsc{rewiring}), and part of this persistence may stem from the context the judge receives.
The failure share therefore characterizes this corpus together with its annotation pipeline.
The inference problem itself does not depend on where label persistence comes from: pooled tests can be affected when both series are dependent, and the human-annotated AnnoMI corpus shows the effect of the sampling unit without an LLM judge.

\paragraph{Methodological assumptions.}
The effective-sample-size correction assumes geometrically decaying (first-order) dependence and does not capture differences between conversations; the block bootstrap does not share these assumptions but treats conversations as independent units.
The simulation study evaluates the criterion on Gaussian AR(1) and binary Markov mechanisms with complete series and conversations as independent units; meeting its pre-specified tolerances is no proof of exact error control, the main procedure was selected by a rule applied to the same simulation, and conversation-specific random effects, the relations among real metrics, and user-level clustering are not simulated (Appendix~\ref{app:simulation}).
With 2{,}000 resamples, BH applied directly to the bootstrap or combined $p$-values of \RNnTests{} tests declares no or at least three discoveries, and the simulated studies with only one or two true associations rarely have them detected (Section~\ref{sec:protocol}).
A post hoc extension of the corpus analysis to \RBtotalB{} resamples per test leaves the \RNrobJoint{} nominal confirmations unchanged and adds one BH discovery (Section~\ref{sec:results}); it shares its first 2{,}000 resamples with the main analysis, and the simulation was not repeated at that budget.
Resampling whole conversations preserves structure that recurs within conversations, including effects of turn position: in post hoc checks, the correlation of one of the \RNbhPfinalJoint{} BH discoveries changed sign in both position checks, and \RCwithinMinor{} of the \RNrobJoint{} nominally confirmed pooled associations retain less than half of their size, or change sign, after centering within conversations (Section~\ref{sec:results}).
The hold-out shares its user population with the exploratory split and was reanalyzed with the main data construction (Section~\ref{sec:setup}).
We do not release the original conversation data due to privacy constraints; the complete per-pair test inventory, the inference code, validation scripts, and synthetic demonstration data are released instead (Appendix~\ref{app:synthetic}).

\appendix

\section{Systematic Literature Survey}
\label{app:survey}

This appendix documents the systematic screen and the full-text audit
summarized in Section~\ref{sec:related}.

\subsection{Protocol}
\label{app:survey-protocol}

\paragraph{Sources and queries.}
Three sources were screened in September 2026:
(i)~the complete ACL Anthology 2020--2026 (\RNaclTotal{} entries, titles
and abstracts);
(ii)~arXiv cs.CL; and
(iii)~OpenReview (ICLR/NeurIPS).
Candidate selection required all three of the following patterns
(case-insensitive regular expressions, documented verbatim in the released
survey protocol):
\begin{flushleft}\small\ttfamily
K1: \detokenize{multi-?turn|dialogu?e|conversation}\\
K2: \detokenize{turn-?level|per-?turn|utterance-?level|}\\
\hspace*{2em}\detokenize{per-?utterance|turn-by-turn}\\
K3: \detokenize{evaluat|benchmark|metric|assess|safety|quality|detect}
\end{flushleft}
The two printed lines of K2 concatenate without a space.

\paragraph{Yield and screening depth.}
The screen yields \RNsurveyCand{} source records: 86 ACL, 171 arXiv,
and 186 OpenReview.  Of these, \RNsurveyFull{} have a full-text screen
(86 ACL and 170 arXiv); \RNsurveyAbs{} have only an abstract-level record
(186 OpenReview and one arXiv record).
Records are not unique works: after normalizing titles, the \RNsurveyCand{}
records form 302 title groups, 217 with at least one full-text screen and
172 with at least one pooled-inference flag; the record-to-group mapping is
released with the supplement.

\paragraph{Screening rules.}
The full-text screen addresses two questions:
(1)~does the paper report inferential statistics (test, $p$-value, confidence
interval, correlation significance) on pooled turn- or utterance-level
data?
(2)~if so, does it handle the serial dependence statistically
(dependence-aware standard errors, effective sample size, per-unit
aggregation before testing)?
The first stage flags a record when turn or utterance vocabulary occurs
within 300 characters of inferential-statistics vocabulary in the extracted
text; the second searches the full text for dependence vocabulary.
The exact patterns and the matching code are released.
These automated flags select passages for inspection.  All
dependence-vocabulary hits were inspected by hand, as was the 30-paper sample
below; the remaining flagged records were not, so the screen measures neither
the precision of the first stage nor how often dependence handling is
described in other terms.
Architectural temporal modeling (Hawkes processes, recurrence over turns,
graph-based dynamics) is categorized separately: modeling time in the
\emph{architecture} does not correct the \emph{statistics} of a pooled
test, and the audit below contains a paper that does both, which makes the
distinction concrete.

\subsection{Results}
\label{app:survey-results}

Of the \RNsurveyFull{} full-text-screened records, \RNsurveyPooled{} meet
the pooled-inference criterion of the first stage and \RNsurveyNoPool{} do
not.
Among the \RNsurveyPooled{}, the dependence-vocabulary stage flags ten
records; hand inspection of all ten identifies \textbf{two}
external papers that handle the dependence statistically:

\begin{itemize}[nosep]
\item \citet{li2025timetoinconsistency} model turn-level robustness decay
  with survival analysis and compute conversation-level cluster-robust
  standard errors---the cluster-robust route;
\item \citet{wu2026sellme} measure the lag-1 autocorrelation of their drift
  quantity (mean 0.099, up to 0.58), report it, and absorb the dependence
  by per-trajectory aggregation before testing---the aggregation route.
\end{itemize}

Of the other eight hits, six use the vocabulary in a different context and
two are not independent external evaluations.

\subsection{Full-text audit of a 30-paper sample}
\label{app:survey-list}

A 30-entry convenience audit records the full-text assessments available for the listed references; the released
audit table gives each entry's identification and classification, including one entry whose bibliographic identity
remains uncertain and the borderline cases of the classification.
The counts below describe this audit record; they are not a validated partition of thirty unambiguously
identified papers.

\begin{itemize}[nosep]
\item \textbf{20 of 30 are out of scope}: they report aggregate scores or
  benchmark accuracies without pooled turn-level inferential statistics
  (among them SODA, LMSYS-Chat-1M, WildChat, Conversation Chronicles,
  MT-Bench-101, Clembench, CoSafe, Crescendo, GOAT, ProsocialDialog, NeMo
  Guardrails).
\item \textbf{5 of 30 pool turn-level tests without any dependence
  handling}, including the largest pooled test in the sample
  (SycEval, \citealp{fanous2025syceval}: 15{,}345 pooled responses with independence-assuming $z$/$\chi^2$
  tests) and ASAP \citep{ye2023modeling}, which models turn dynamics with a Hawkes process and
  still evaluates with a pooled paired $t$-test---the concrete case
  separating architectural temporal modeling from statistical correction.
\item \textbf{4 of 30 handle it partially} (e.g., conversation-level
  inference next to turn-share tests treated as independent Bernoulli
  draws; turn-level rank correlations reported without $p$-values).
\item \textbf{1 of 30 addresses it}, by per-person aggregation before
  testing with the independence question explicitly discussed (Utterance
  Emotion Dynamics; \citealp{teodorescu2023language})---the same aggregation route that
  \citet{wu2026sellme} take.
\end{itemize}

\subsection{Scope of the evidence}
\label{app:survey-claim}

The screen and the audit support three statements: pooled turn-level tests
without dependence handling occur in the audited literature, including tests
over thousands of responses; statistical dependence handling exists and is
compatible with current practice (three audited or inspected examples); and
the automated screen surfaces few papers that describe such handling.
They do not provide an estimate of how prevalent dependence handling is in
the field.
The complete record table (one row per source record with source,
bibliographic key, screening flags, logged evidence excerpt where available,
and screening depth) ships in the supplement.

\section{Full Results and Sensitivity Constructions}
\label{app:results}

Table~\ref{tab:full} lists all \RNpooledRaw{} metric--label pairs that are
significant under raw pooled testing, with both autocorrelation profiles and
the outcome of the dependence-aware procedure.  The machine-readable inventory
of all \RNnTests{} tests, with every intermediate quantity and the per-pair
flags of the user-sensitivity and hold-out analyses, ships in the supplement
together with validation scripts for the reported tables.

\paragraph{Criterion by analysis.}
The full-corpus and the public-corpus results use the criterion
$p_{\mathrm{final}}$ of Eq.~\ref{eq:robust-criterion} in its joint form with
the bootstrap value of Eq.~\ref{eq:p-boot}.  The user-sensitivity analyses and
the exploratory groups and strict replication criterion of the hold-out were
run with the conservative-bound form (Eq.~\ref{eq:chelton} in place of
Eq.~\ref{eq:chelton-joint}); on the full corpus both forms give the same
\RNrobJoint{} robust pairs.  The six directional hypotheses use their
pre-registered test (Spearman correlation, one-sided conversation-level
bootstrap, Holm).  The AnnoMI analysis is an interval comparison
(Section~\ref{sec:annomi-public}).  The simulation evaluates the joint form,
the conservative-bound form, and the bootstrap alone on the same simulated
data (Appendix~\ref{app:simulation}).

{\scriptsize\setlength{\tabcolsep}{3pt}
\begin{longtable}{@{}lllrrrrrlcc@{}}
\caption{All \RNpooledRaw{} metric--label pairs that are significant under raw pooled testing (main construction). $\bar\rho_x$, $\bar\rho_y$: mean per-conversation lag-1 autocorrelation of metric and label on the test's turn pairs; $n$: available turn pairs of the test; $n_{\mathrm{eff}}$: joint effective sample size (Eq.~\ref{eq:chelton-joint}); $p_{\mathrm{boot}}$: conversation-level bootstrap (Eq.~\ref{eq:p-boot}), $\dagger$ marks tests without any opposite-sign replicate among the 2{,}000 resamples; Rob.: $p_{\mathrm{final}}<0.05$; BH: discovery under Benjamini--Hochberg on $p_{\mathrm{final}}$; $\ast$: robust pair whose sign changes in at least one leave-one-user-out run (Section~\ref{sec:users}).}\label{tab:full}\\
\toprule
Label & Metric & Category & $r$ & $\bar\rho_x$ & $\bar\rho_y$ & $n$ & $n_{\mathrm{eff}}$ & $p_{\mathrm{boot}}$ & Rob. & BH \\
\midrule
\endfirsthead
\toprule
Label & Metric & Category & $r$ & $\bar\rho_x$ & $\bar\rho_y$ & $n$ & $n_{\mathrm{eff}}$ & $p_{\mathrm{boot}}$ & Rob. & BH \\
\midrule
\endhead
\bottomrule
\endfoot
malicious & det\_B\_t & Cumulative accum. & 0.206 & 0.898 & 0.298 & 11{,}100 & 6{,}412 & 0.0320 & yes & no \\
malicious & Toughness & Cumulative accum. & 0.186 & 0.922 & 0.315 & 11{,}395 & 6{,}260 & 0.0590 & no & no \\
rewiring & drift\_score & Frame distance & -0.061 & 0.587 & 0.133 & 11{,}240 & 9{,}609 & 0.0010$\dagger$ & yes & yes \\
rewiring & WD & Frame distance & -0.087 & 0.683 & 0.116 & 10{,}670 & 9{,}105 & 0.0010$\dagger$ & yes & yes \\
malicious & LVSK & Lexical/structural & -0.036 & 0.405 & 0.316 & 11{,}391 & 8{,}806 & 0.0800 & no & no \\
rewiring & LVSK & Lexical/structural & -0.049 & 0.406 & 0.138 & 11{,}428 & 10{,}213 & 0.0010$\dagger$ & yes & yes \\
malicious & MAS & Lexical/structural & 0.110 & 0.178 & 0.337 & 11{,}574 & 10{,}263 & 0.0170 & yes & no \\
malicious & E\_Modul & Thermodynamic & 0.042 & 0.665 & 0.222 & 7{,}912 & 5{,}873 & 0.2199 & no & no \\
malicious & Entropy & Thermodynamic & 0.027 & 0.582 & 0.315 & 11{,}395 & 7{,}861 & 0.1309 & no & no \\
malicious & Psi & Thermodynamic & -0.027 & 0.582 & 0.315 & 11{,}395 & 7{,}861 & 0.1149 & no & no \\
malicious & Hysteresis\_rolling & Thermodynamic & -0.149 & 0.696 & 0.235 & 9{,}492 & 6{,}826 & 0.0010$\dagger$ & yes & yes \\
typx & PRO\_rolling & Rolling aggr. (mixed W) & 0.043 & 0.761 & 0.284 & 11{,}607 & 7{,}478 & 0.0620 & no & no \\
typx & det\_B\_t & Cumulative accum. & 0.026 & 0.900 & 0.275 & 11{,}133 & 6{,}716 & 0.6077 & no & no \\
typx & DIV & Embedding velocity & 0.025 & 0.040 & 0.284 & 11{,}607 & 11{,}348 & 0.2419 & no & no \\
typx & QR & Lexical/structural & -0.038 & 0.082 & 0.284 & 11{,}607 & 11{,}077 & 0.0350 & yes & no \\
rewiring & FID & Frame distance & -0.044 & 0.191 & 0.124 & 10{,}854 & 10{,}351 & 0.0710 & no & no \\
rewiring & NC2 & Lexical/structural & 0.084 & 0.890 & 0.124 & 10{,}854 & 8{,}691 & 0.0010$\dagger$ & yes & yes \\
malicious & MR & Lexical/structural & 0.023 & 0.088 & 0.337 & 11{,}574 & 10{,}909 & 0.3978 & no & no \\
malicious & F\_CV & Lexical/structural & -0.050 & 0.554 & 0.275 & 10{,}447 & 7{,}681 & 0.0130 & yes & yes \\
malicious & det\_c\_t & Impulse/indicator & 0.106 & 0.307 & 0.298 & 11{,}100 & 9{,}239 & 0.0010$\dagger$ & yes & yes \\
malicious & B\_windowed & Rolling aggr. (mixed W) & 0.156 & 0.666 & 0.241 & 7{,}681 & 5{,}558 & 0.0020 & yes & yes \\
malicious & Toughness\_rate & Normalized cumul. & -0.036 & 0.763 & 0.315 & 11{,}395 & 6{,}975 & 0.3438 & no & no \\
typx & det\_c\_t & Impulse/indicator & 0.063 & 0.306 & 0.275 & 11{,}133 & 9{,}406 & 0.0010$\dagger$ & yes & yes \\
malicious & dir\_consistency & Directional consist. & -0.056 & -0.014 & 0.304 & 11{,}166 & 11{,}262 & 0.0010$\dagger$ & yes & yes \\
malicious & sem\_velocity$\ast$ & Embedding velocity & 0.048 & 0.091 & 0.315 & 11{,}395 & 10{,}759 & 0.0020 & yes & yes \\
rewiring & dir\_consistency & Directional consist. & -0.027 & -0.013 & 0.133 & 11{,}203 & 11{,}243 & 0.0010$\dagger$ & yes & yes \\
typx & hooke\_residual$\ast$ & Thermodynamic & -0.050 & 0.079 & 0.281 & 10{,}185 & 9{,}742 & 0.0220 & yes & no \\
malicious & epsilon\_wd\_rel & Thermodynamic & 0.059 & 0.682 & 0.283 & 10{,}635 & 7{,}193 & 0.1159 & no & no \\
typx & nyp\_mech & Thermodynamic & 0.091 & 0.447 & 0.305 & 3{,}483 & 2{,}648 & 0.1049 & no & no \\
malicious & coherence\_trap & Interaction & 0.036 & 0.063 & 0.287 & 10{,}818 & 10{,}432 & 0.1279 & no & no \\
malicious & L\_weighted & Weighted variants & -0.040 & 0.501 & 0.304 & 11{,}203 & 8{,}239 & 0.0740 & no & no \\
malicious & WD\_weighted & Weighted variants & -0.080 & 0.669 & 0.283 & 10{,}635 & 7{,}249 & 0.0060 & yes & yes \\
rewiring & WD\_weighted & Weighted variants & -0.086 & 0.670 & 0.116 & 10{,}670 & 9{,}133 & 0.0010$\dagger$ & yes & yes \\
rewiring & FID\_weighted & Frame distance & -0.037 & 0.183 & 0.124 & 10{,}854 & 10{,}371 & 0.1359 & no & no \\
malicious & coherence\_trap\_weighted & Interaction & 0.037 & 0.061 & 0.287 & 10{,}818 & 10{,}447 & 0.1249 & no & no \\
malicious & NCD & Compression (NCD) & 0.047 & 0.106 & 0.337 & 11{,}574 & 10{,}772 & 0.1279 & no & no \\
typx & NCD$\ast$ & Compression (NCD) & 0.131 & 0.107 & 0.284 & 11{,}607 & 10{,}922 & 0.0010$\dagger$ & yes & yes \\
typx & sem\_velocity\_w$\ast$ & Embedding velocity & -0.103 & 0.398 & 0.280 & 11{,}428 & 9{,}140 & 0.0010$\dagger$ & yes & yes \\
malicious & delta\_FID & Impulse/indicator & -0.041 & -0.352 & 0.283 & 10{,}635 & 12{,}990 & 0.0010$\dagger$ & yes & yes \\
malicious & delta\_NC2 & Impulse/indicator & -0.127 & 0.640 & 0.283 & 10{,}635 & 7{,}369 & 0.0020 & yes & yes \\
malicious & delta\_NCD & Impulse/indicator & 0.028 & -0.393 & 0.315 & 11{,}395 & 14{,}614 & 0.0010$\dagger$ & yes & yes \\
typx & delta\_FID & Impulse/indicator & -0.026 & -0.353 & 0.272 & 10{,}666 & 12{,}931 & 0.0010$\dagger$ & yes & yes \\
rewiring & delta\_FID & Impulse/indicator & -0.038 & -0.353 & 0.116 & 10{,}670 & 11{,}579 & 0.0010$\dagger$ & yes & yes \\
typx & E\_ratio & Thermodynamic & 0.033 & 0.231 & 0.240 & 9{,}494 & 8{,}495 & 0.2049 & no & no \\
malicious & WD\_x\_FCV & Interaction & -0.029 & 0.562 & 0.275 & 10{,}447 & 7{,}646 & 0.2519 & no & no \\
malicious & WD\_div\_FCV & Interaction & -0.032 & 0.549 & 0.275 & 10{,}447 & 7{,}704 & 0.1999 & no & no \\
malicious & NCD\_x\_DIV & Interaction & 0.019 & 0.043 & 0.337 & 11{,}574 & 11{,}246 & 0.3028 & no & no \\
typx & NCD\_x\_DIV$\ast$ & Interaction & 0.061 & 0.043 & 0.284 & 11{,}607 & 11{,}327 & 0.0130 & yes & yes \\
malicious & sem\_velocity\_t$\ast$ & Timestamp/recovery & 0.134 & 0.129 & 0.311 & 9{,}583 & 8{,}847 & 0.0010$\dagger$ & yes & yes \\
typx & sem\_velocity\_t$\ast$ & Timestamp/recovery & -0.071 & 0.128 & 0.276 & 9{,}616 & 8{,}960 & 0.0120 & yes & yes \\
malicious & turn\_pressure$\ast$ & Timestamp/recovery & 0.092 & 0.351 & 0.311 & 9{,}583 & 7{,}700 & 0.0010$\dagger$ & yes & yes \\
typx & turn\_pressure & Timestamp/recovery & -0.035 & 0.351 & 0.276 & 9{,}616 & 7{,}919 & 0.1779 & no & no \\
malicious & EWMA\_sem\_velocity & Embedding velocity & 0.093 & 0.653 & 0.315 & 11{,}395 & 7{,}503 & 0.0010$\dagger$ & yes & yes \\
malicious & wd\_oscillation & Thermodynamic & -0.053 & 0.655 & 0.245 & 8{,}938 & 6{,}460 & 0.0230 & yes & no \\
typx & E\_t & Thermodynamic & -0.021 & -0.069 & 0.257 & 9{,}935 & 10{,}294 & 0.6557 & no & no \\
malicious & sigma\_user & Embedding velocity & -0.049 & 0.008 & 0.283 & 10{,}635 & 10{,}588 & 0.0010$\dagger$ & yes & yes \\
malicious & sigma\_user\_base & Embedding velocity & -0.024 & 0.016 & 0.283 & 10{,}635 & 10{,}539 & 0.3198 & no & no \\
malicious & epsilon\_ai & Embedding velocity & -0.033 & 0.170 & 0.283 & 10{,}635 & 9{,}656 & 0.4878 & no & no \\
malicious & E\_ai & Embedding velocity & -0.032 & 0.063 & 0.283 & 10{,}635 & 10{,}263 & 0.3128 & no & no \\
typx & sigma\_user & Embedding velocity & 0.024 & 0.008 & 0.272 & 10{,}666 & 10{,}622 & 0.3468 & no & no \\
typx & epsilon\_ai & Embedding velocity & -0.070 & 0.171 & 0.272 & 10{,}666 & 9{,}720 & 0.1000 & no & no \\
typx & E\_ai & Embedding velocity & 0.042 & 0.063 & 0.272 & 10{,}666 & 10{,}306 & 0.3518 & no & no \\
typx & sv\_user & Embedding velocity & 0.035 & 0.080 & 0.280 & 11{,}428 & 10{,}931 & 0.0110 & yes & yes \\
typx & sv\_ai & Embedding velocity & -0.028 & 0.154 & 0.280 & 11{,}428 & 10{,}482 & 0.1759 & no & no \\
typx & asymm & Embedding velocity & -0.029 & 0.094 & 0.280 & 11{,}428 & 10{,}842 & 0.0810 & no & no \\
rewiring & sigma\_user & Embedding velocity & -0.036 & 0.009 & 0.116 & 10{,}670 & 10{,}647 & 0.0020 & yes & yes \\
rewiring & epsilon\_ai & Embedding velocity & -0.070 & 0.172 & 0.116 & 10{,}670 & 10{,}254 & 0.0030 & yes & yes \\
\end{longtable}
}

\paragraph{Sensitivity constructions.}
The main construction (H) treats metric values that are undefined at the
start of a conversation as missing, propagates this missingness into dependent
differences, products, and windows, and rebuilds recursive, cumulative, and
fitted quantities from their measured inputs (Section~\ref{sec:setup}).
Three alternatives were specified together with it and are computed on the
same data.
Construction A keeps the initialization values and numerical fallbacks as
computed.
N1 applies the missing-value and eligibility rules of H without rebuilding
the recursive, cumulative, and fitted quantities.
N2 restricts the missing-value treatment to the frame-distance metrics and
the quantities that depend on them.
Table~\ref{tab:variants} gives the funnel for each construction, and for H
without the four tests on label-fitted metrics (Appendix~\ref{app:metrics}).

\begin{table}[!ht]
\centering\small
\begin{tabular}{@{}lrrrrrr@{}}
\toprule
Construction & Tests & Raw pooled & BH screen & Robust & FS (\%) & BH on $p_{\mathrm{final}}$ \\
\midrule
A & 113 & 86 & 83 & 53 & 36.1 & 49 \\
H (main) & 113 & 67 & 65 & 35 & 46.2 & 30 \\
N1 & 113 & 67 & 64 & 37 & 42.2 & 32 \\
N2 & 113 & 83 & 79 & 50 & 36.7 & 47 \\
H without label-fitted tests & 109 & 65 & 63 & 34 & 46.0 & 30 \\
\bottomrule
\end{tabular}
\caption{Correction funnel per data construction.  Robust: BH-screened pairs
with $p_{\mathrm{final}}<0.05$; FS: failure share
(Eq.~\ref{eq:inflation-rate}).}
\label{tab:variants}
\end{table}

\begin{table}[!ht]
\centering\small
\begin{tabular}{@{}llrrrrrr@{}}
\toprule
 & & & & \multicolumn{2}{c}{Strict} & \multicolumn{2}{c}{Bootstrap leg only} \\
Constr. & Tier & $N_R$ & $N_P$ & $R$ & $P$ & $R$ & $P$ \\
\midrule
A & A & 48 & 23 & 16 & 2 & 23 & 5 \\
A & B & 48 & 23 & 20 & 3 & 28 & 6 \\
H & A & 35 & 21 & 10 & 2 & 16 & 4 \\
H & B & 35 & 21 & 13 & 1 & 21 & 3 \\
N1 & A & 36 & 20 & 10 & 2 & 17 & 4 \\
N1 & B & 36 & 20 & 13 & 1 & 22 & 3 \\
N2 & A & 46 & 23 & 15 & 3 & 22 & 6 \\
N2 & B & 46 & 23 & 19 & 3 & 27 & 6 \\
\bottomrule
\end{tabular}
\caption{Hold-out replication counts in all four constructions.  $N_R$ and
$N_P$: dependence-aware and pooled-only findings in the exploratory set;
$R$ and $P$: number of them that replicate under the strict and the
bootstrap-only criterion (Section~\ref{sec:holdout}).}
\label{tab:holdout-all}
\end{table}

\paragraph{The six pre-registered hypotheses.}
The pre-registered directional hypotheses are, in their original order:
H1 \textsc{malicious}--\texttt{B\_windowed}, positive;
H2 \textsc{malicious}--\texttt{det\_B\_t}, positive;
H3 \textsc{malicious}--\texttt{EWMA\_sv} (\texttt{EWMA\_sem\_velocity} in Table~\ref{tab:full}), positive;
H4 \textsc{malicious}--\texttt{delta\_NC2}, negative;
H5 \textsc{typx}--\texttt{NCD}, positive;
H6 \textsc{typx}--\texttt{det\_c\_t}, positive.
Each is tested with a Spearman correlation and a one-sided
whole-conversation bootstrap with 10{,}000 resamples, followed by Holm's
step-down procedure at family-wise $\alpha=0.05$.
In the main construction, H3 and H5 are confirmed in Tier~A, and H5 and H6
in Tier~B; N1 gives the same result.
Constructions A and N2 confirm three hypotheses in Tier~A (H3, H5, H6) and
four in Tier~B (H1, H3, H5, H6).
H4, which is not confirmed in either tier of any construction, rests on the
first difference of a windowed quantity; on the full corpus its pooled
association with the label depends on turn position
(Section~\ref{sec:results}).
The pre-registration defines two or three confirmed hypotheses together with
a met meta-level criterion as partial confirmation, and four or more as
confirmation.

\section{Annotation Protocol}
\label{app:annotation}

Turn-level annotations were produced by a structured LLM-as-judge pipeline whose documented default model is Claude Opus 4.6 \citep{anthropic2026opus46}.
The protocol follows a three-stage design:

\paragraph{Stage 1: Arc Analysis.}
The complete conversation is submitted to the LLM with instructions to identify the overall narrative structure, turning points, and phases of escalation or de-escalation.

\paragraph{Stage 2: Batched Turn-by-Turn Annotation.}
The conversation is processed in batches of 30 turns; each later batch receives the preceding five annotation rows as context.
Each turn receives binary labels for three dimensions: \textsc{malicious} (manipulative assistant behavior), \textsc{typx} (emergent co-construction), and \textsc{rewiring} (semantic redefinition of established terms).

\paragraph{Stage 3: Consistency Check.}
An optional consistency pass checks the combined annotations; where it returns no usable result, the Stage-2 annotations are kept.

The annotation schema produces 17 columns per turn, of which 3 binary labels are used for the statistical analyses in this paper.

\paragraph{Expert review.}
An expert review of the annotation was conducted by a computational linguist (specializing in computational linguistics and German studies). The reviewer examined a stratified sample of annotated conversations covering the three manipulation categories of the annotation codebook (semantic rewiring, compliance drift, and elicitation) and assessed whether the assigned labels were consistent with the operational definitions in the annotation codebook. The reviewer confirmed the annotation quality and flagged no systematic disagreements. This qualitative review supports the linguistic interpretation of the labels; it is not a formal inter-rater reliability study.

\paragraph{Label autocorrelation.}
\label{sec:labelrho}
The label series are serially dependent: median per-conversation lag-1 autocorrelations are \RNrhoYmal{} (\textsc{malicious}), \RNrhoYtypx{} (\textsc{typx}), and \RNrhoYrw{} (\textsc{rewiring}).
This dependence is an input to the corrected inference (Section~\ref{sec:chelton}), where it is estimated on the turn pairs available for each test, and it is reported per pair in the released inventory.

\paragraph{Batch boundaries.}
We examined whether label switches were more frequent at recorded partition boundaries than at interior positions.
For two recorded partitions of the source files, the switch rate at partition boundaries relative to interior positions is
\RNchunkAMal/\RNchunkATypx/\RNchunkARw{} and
\RNchunkBMal/\RNchunkBTypx/\RNchunkBRw{} for \textsc{malicious}/\textsc{typx}/\textsc{rewiring},
and a per-conversation rotation test finds boundary alignment in
\RNchunkAMalPSig{} of \RNchunkAMalPCount{},
\RNchunkATypxPSig{} of \RNchunkATypxPCount{}, and
\RNchunkARwPSig{} of \RNchunkARwPCount{} testable conversations, respectively, under both partitions (one conversation without complete position records is excluded).
For \textsc{typx} and \textsc{rewiring}, switches are thus rarer at the recorded boundaries than inside; for \textsc{malicious} they are slightly more frequent.
The recorded partitions need not coincide with the judge's processing batches, and the context passed between batches (Stage~2) remains a possible contributor to label persistence; the analysis does not separate it from the persistence of the annotated behavior.
On the public corpora with human per-utterance labels, persistence varies with the labeled construct (Section~\ref{sec:publicdata}).

\paragraph{Role of the annotation source in the correction.}
\label{sec:agnostic}
The correction methodology applies regardless of where the labels come from (human, LLM judge, or automatic classifier): the procedure estimates both series' autocorrelations from the data at hand and resamples whole conversations.
What the annotation source does affect is the amount of label dependence and hence how much pooled inference overstates the evidence.
The simulation study (Appendix~\ref{app:simulation}) quantifies this: where either series is serially independent, the pooled test stays near its nominal level (\SKPoolIndLo--\SKPoolIndHi), and its rejection rate grows with the dependence of both series, reaching \SKPoolMaxA{} at the upper end of the simulated grid.

\section{Synthetic Demonstration Data}
\label{app:synthetic}

To show how the released code is run without private conversation data, the supplement contains three synthetic data sets of 100 conversations each, produced with the generators of the simulation study (Appendix~\ref{app:simulation}):

\begin{enumerate}[nosep]
\item Conversation lengths are drawn once, uniformly from 30 to 500 turns.
\item The label has a base rate of 0.25 and a population lag-1 autocorrelation of 0.6; it is generated (a) by thresholding a Gaussian AR(1) latent series or (b) by a two-state Markov chain.
\item The metric has a population lag-1 autocorrelation of 0.7. Two data sets carry no association, one per label mechanism. In the third, under mechanism~(a), the metric is a mixture of the latent label series and an independent AR(1) series, with a population correlation of 0.10 with the label; it is then not itself an AR(1) process.
\item Each data set is analyzed with the pooled test, both analytic corrections, the conversation-level bootstrap, both forms of the combined criterion, and the cluster-robust reference; the stored output lists all $p$-values.
\end{enumerate}

The demonstration data are single draws that illustrate how the code is run; they are no calibration study, and the rejection and detection rates of this paper come from the simulation study of Appendix~\ref{app:simulation}.
The data sets, their generation script, and the simulation code accompany the journal submission as supplementary material.

\section{Simulation Study: Rejection Rates and Detection under Controlled Dependence}
\label{app:simulation}

This appendix evaluates the components of the testing protocol on simulated
data with known truth, for single tests and for whole studies.
The design, the procedures, the replication numbers, and the evaluation rule
were fixed before the run, and the run was executed once.

\paragraph{Design.}
Each simulated data set has the size of the corpus: $k = \RNnConv$
conversations with the empirical lengths ($n = \RNnPairs$ turn pairs) and
the base rate of the \textsc{malicious} label (0.244).
Within a conversation every series is stationary from the first turn.
Label series are generated by two mechanisms: (a)~thresholding a
unit-variance Gaussian AR(1) latent series $U_t$ at the value that gives the
base rate, $B_t=\mathbf{1}\{U_t>\tau\}$ with $\tau=0.694$, and (b)~a
two-state Markov chain with the same base rate.
The metric is $X_t = g\,U_t+\sqrt{1-g^2}\,Z_t$ with an independent Gaussian
AR(1) series $Z_t$; without an association $g=0$, and associations are
simulated under mechanism~(a).
All generator settings are determined from population targets: $g$ from the
target correlation $r$ between $X_t$ and $B_t$ and the coefficient of $Z_t$
from the target lag-1 autocorrelation $\rho_x$ of the metric algebraically,
and the coefficient of $U_t$ from the target lag-1 autocorrelation $\rho_y$
of the label by deterministic numerical inversion; no setting is tuned on
simulated samples.
The null grid crosses $\rho_x\in\{0,0.2,0.5,0.7,0.9,0.95\}$ with
$\rho_y\in\{0,0.35,0.6,0.9,0.97\}$ under both mechanisms (60 cells); the grid
with associations crosses $r\in\{0.05,0.10,0.20\}$,
$\rho_x\in\{0.2,0.7,0.9\}$, and $\rho_y\in\{0,0.35,0.9,0.97\}$ (36 cells).
At equal lag-1 value the two mechanisms differ in memory: under~(a) the label
autocorrelation at lag $h$ decays more slowly than $\rho_y^{\,h}$ (for
$\rho_y=0.9$ it is 0.690 at lag 10 and 0.366 at lag 50, against 0.349 and
0.005 under~(b)).
With an association the metric is a mixture of two AR(1) series: its lag-1
value is set, and its autocorrelation function is not that of an AR(1)
process.
The targets are population values.  The mean per-conversation estimates that
enter the analytic correction are lower in these finite series, for both
series in every null cell: for the metric, targets of 0.5, 0.9, and 0.95
correspond to mean
estimates of \SKrxHatMid, \SKrxHatHigh, and \SKrxHatTop, and the estimates for
the label are listed in Table~\ref{tab:simnull}.
Every data set is analyzed by all procedures on the same series and the same
$B=2{,}000$ conversation-level resamples: the pooled test; the joint and the
conservative-bound analytic correction alone; the bootstrap alone
(Eq.~\ref{eq:p-boot}); the combined criterion $p_{\mathrm{final}}$ in its
joint and its conservative-bound form; and, as a textbook reference, a
$t$-test of the regression slope with standard errors clustered by
conversation \citep{cameron2015practitioner}.
Null cells have \SKnullReps{} replications, cells with an association
\SKpowReps.

\paragraph{Study-level conditions.}
A simulated study has \RNnTests{} tests with the lag-1 values $\rho_x$ of the
battery, including its \RNnegTests{} negative ones.
In eight conditions the tests share three label series, used by 59, 35, and
19 tests as in the corpus, with label persistence $\rho_y=0.35$ or $0.60$ and
a probability of 0\%, 5\%, 20\%, or 40\% with which each test carries a true
association ($r=0.10$); the number of true associations therefore varies
between studies (from \SDtrueMin{} to \SDtrueMax{} at 5\%).
The two persistence levels are simulation settings chosen at the scale of
the observed estimates, not an estimated interval for the corpus.
Two further conditions generate a separate label series for every test
($\rho_y=0.35$; 20\% or 40\%).
Each condition has \SKstudies{} studies.
Metrics with a true association to the same label share its latent series
and are therefore weakly correlated; without true associations the metrics
are mutually independent.
The algebraic relations among real metrics and the reuse of a metric across
labels are not reproduced.

\paragraph{Evaluation rule.}
Two conditions were specified in advance.
(N)~In no null cell does the observed rejection rate at $\alpha=0.05$ exceed
an upper tolerance of \SKtol, one and a half times the nominal level; this
is a robustness tolerance and not a demonstration of an exact 5\% test.
(S)~In every study condition the observed mean false discovery proportion
under BH on the procedure's $p$-values is at most 5\%; this is a rule on a
Monte Carlo estimate and not a guaranteed control.
The joint form of the combined criterion was the first candidate for the
main procedure and the conservative-bound form the second.
Selecting between them by this rule is a selection on the same simulation,
not an independent confirmation of the selected procedure.

\paragraph{Rejection rates without association.}
Table~\ref{tab:simnull} reports the null rejection rate per cell for the
pooled test and for the combined criterion in its joint form.
\begin{itemize}[nosep]
\item \textbf{Pooled test.}  Where one of the two series is serially
  independent ($\rho_x=0$ or $\rho_y=0$; 20 cells), its rejection rate lies
  between \SKPoolIndLo{} and \SKPoolIndHi.  Where both are dependent it
  exceeds \SKtol{} in 19 of 20 cells under each mechanism and grows with both
  dependences, up to \SKPoolMaxA{} under mechanism~(a) and \SKPoolMaxB{}
  under mechanism~(b); at $\rho_y=0.35$, the level whose realized label
  estimates are nearest to those observed in the corpus, it reaches
  \SKPoolCorpA{} and \SKPoolCorpB.
\item \textbf{Combined criterion, joint form.}  Its rejection rates lie
  between \SKJointMin{} and \SKJointMax{} over all 60 cells (pointwise 95\%
  Wilson interval of the largest: \SKJointMaxLo--\SKJointMaxHi); no cell
  exceeds the tolerance.  It is not an exact 5\% test: \SKJointAtLimit{}
  cells reach at least \SKmcLimitCount{} rejections out of \SKnullReps, which
  an exact test would rarely produce even allowing for the 60 cells of one
  procedure.
\item \textbf{Other procedures.}  The conservative-bound form ranges from
  \SKConsMin{} to \SKConsMax{} and falls below 0.025 in \SKConsBelow{} cells;
  the bootstrap alone ranges from \SKBootMin{} to \SKBootMax{} and the
  cluster-robust reference from \SKCRMin{} to \SKCRMax.
  The analytic corrections alone do not meet the tolerance: the joint
  correction reaches \SKFJointMax{} and exceeds \SKtol{} in \SKFJointOver{}
  cells, the conservative bound reaches \SKFConsMax{} (\SKFConsOver{} cells);
  in these cells the combined criterion is carried by the bootstrap.
  Averaging the label autocorrelation over all conversations, with constant
  sequences counted as zero, leaves the combined criterion nearly unchanged
  (largest rate \SKJointAllMax).
\end{itemize}
With \SKnullReps{} replications the Monte Carlo standard error at a true
rate of 0.05 is \SKmcSE.

\begin{table}[t]
\centering
\footnotesize
\begin{tabular}{@{}lccccccc@{}}
\toprule
$\rho_y$ & $\hat\rho_y$ & $\rho_x{=}0$ & $\rho_x{=}0.2$ & $\rho_x{=}0.5$ & $\rho_x{=}0.7$ & $\rho_x{=}0.9$ & $\rho_x{=}0.95$ \\
\midrule
\multicolumn{8}{@{}l}{\emph{Mechanism (a), thresholded latent series: pooled test}}\\
0    & $-$0.035 & 0.054 & 0.050 & 0.047 & 0.051 & 0.049 & 0.051 \\
0.35 & 0.270 & 0.050 & 0.061 & 0.103 & 0.137 & 0.181 & 0.201 \\
0.6  & 0.452 & 0.047 & 0.084 & 0.157 & 0.236 & 0.360 & 0.390 \\
0.9  & 0.597 & 0.057 & 0.095 & 0.238 & 0.368 & 0.578 & 0.654 \\
0.97 & 0.613 & 0.052 & 0.104 & 0.240 & 0.394 & 0.613 & 0.706 \\
\midrule
\multicolumn{8}{@{}l}{\emph{Mechanism (a), thresholded latent series: combined criterion (joint form)}}\\
0    & $-$0.035 & 0.047 & 0.045 & 0.044 & 0.048 & 0.046 & 0.048 \\
0.35 & 0.270 & 0.044 & 0.042 & 0.050 & 0.054 & 0.053 & 0.054 \\
0.6  & 0.452 & 0.043 & 0.051 & 0.057 & 0.055 & 0.056 & 0.053 \\
0.9  & 0.597 & 0.054 & 0.047 & 0.062 & 0.056 & 0.058 & 0.062 \\
0.97 & 0.613 & 0.046 & 0.052 & 0.060 & 0.054 & 0.059 & 0.059 \\
\midrule
\multicolumn{8}{@{}l}{\emph{Mechanism (b), Markov chain: pooled test}}\\
0    & $-$0.034 & 0.048 & 0.050 & 0.047 & 0.056 & 0.049 & 0.052 \\
0.35 & 0.280 & 0.049 & 0.072 & 0.107 & 0.125 & 0.162 & 0.159 \\
0.6  & 0.498 & 0.054 & 0.086 & 0.144 & 0.203 & 0.271 & 0.304 \\
0.9  & 0.758 & 0.053 & 0.101 & 0.225 & 0.325 & 0.488 & 0.549 \\
0.97 & 0.830 & 0.052 & 0.102 & 0.254 & 0.377 & 0.583 & 0.661 \\
\midrule
\multicolumn{8}{@{}l}{\emph{Mechanism (b), Markov chain: combined criterion (joint form)}}\\
0    & $-$0.034 & 0.043 & 0.046 & 0.042 & 0.049 & 0.045 & 0.048 \\
0.35 & 0.280 & 0.046 & 0.048 & 0.048 & 0.051 & 0.056 & 0.049 \\
0.6  & 0.498 & 0.050 & 0.052 & 0.052 & 0.056 & 0.054 & 0.055 \\
0.9  & 0.758 & 0.050 & 0.050 & 0.054 & 0.049 & 0.049 & 0.055 \\
0.97 & 0.830 & 0.049 & 0.043 & 0.055 & 0.060 & 0.056 & 0.052 \\
\bottomrule
\end{tabular}
\caption{Null rejection rates at $\alpha = 0.05$ (\SKnullReps{} replications
per cell) for the pooled test and the combined criterion
$\max(p_{\mathrm{Chelton}}^{\mathrm{joint}}, p_{\mathrm{boot}})$ under both
label mechanisms.  Rows give the population lag-1 autocorrelation $\rho_y$ of
the label and the realized mean per-conversation estimate $\hat\rho_y$ over
the conversations in which the label varies (shown for $\rho_x = 0.5$).}
\label{tab:simnull}
\end{table}

\paragraph{Detection with association.}
Table~\ref{tab:simpower} reports detection rates under mechanism~(a).
The realized mean correlations lie within \SKrHatMaxDev{} of their targets
(\SKrHatFiveLo--\SKrHatFiveHi, \SKrHatTenLo--\SKrHatTenHi, and
\SKrHatTwentyLo--\SKrHatTwentyHi) and are listed per cell.
At $r=0.20$ the combined criterion detects the association in all but one of
the \SKpowTwentyReps{} replications.
At $r=0.10$ its detection rate is at least 0.95 in \SKpowTenAtLeast{} of 12
cells and falls to \SKpowJointTenMin{} at $(\rho_x,\rho_y)=(0.9,0.97)$.
At $r=0.05$ it ranges from \SKpowJointFiveMin{} to \SKpowJointFiveMax.
Detection is lower in the cells with weak effects and strong dependence of
both series.  The pooled test detects more often there, but in the null
cells with the same dependence ($\rho_x\in\{0.7,0.9\}$, $\rho_y\geq0.9$) it
rejects at rates between \SKPoolStrongLo{} and \SKPoolStrongHi{} without any
association (Table~\ref{tab:simnull}).

\begin{table}[t]
\centering
\footnotesize
\begin{tabular}{@{}cc ccc ccc ccc@{}}
\toprule
 & & \multicolumn{3}{c}{$r=0.05$} & \multicolumn{3}{c}{$r=0.10$} & \multicolumn{3}{c}{$r=0.20$} \\
$\rho_x$ & $\rho_y$ & $\hat r$ & pooled & comb. & $\hat r$ & pooled & comb. & $\hat r$ & pooled & comb. \\
\midrule
0.2 & 0 & 0.050 & 1.000 & 1.000 & 0.100 & 1.000 & 1.000 & 0.200 & 1.000 & 1.000 \\
0.2 & 0.35 & 0.050 & 1.000 & 1.000 & 0.100 & 1.000 & 1.000 & 0.200 & 1.000 & 1.000 \\
0.2 & 0.9 & 0.050 & 0.998 & 0.991 & 0.099 & 1.000 & 1.000 & 0.199 & 1.000 & 1.000 \\
0.2 & 0.97 & 0.050 & 0.995 & 0.983 & 0.099 & 1.000 & 1.000 & 0.199 & 1.000 & 1.000 \\
0.7 & 0 & 0.050 & 1.000 & 1.000 & 0.100 & 1.000 & 1.000 & 0.200 & 1.000 & 1.000 \\
0.7 & 0.35 & 0.050 & 0.994 & 0.980 & 0.101 & 1.000 & 1.000 & 0.200 & 1.000 & 1.000 \\
0.7 & 0.9 & 0.050 & 0.943 & 0.700 & 0.099 & 1.000 & 0.999 & 0.198 & 1.000 & 1.000 \\
0.7 & 0.97 & 0.050 & 0.930 & 0.646 & 0.099 & 0.999 & 0.993 & 0.198 & 1.000 & 1.000 \\
0.9 & 0 & 0.050 & 1.000 & 1.000 & 0.100 & 1.000 & 1.000 & 0.201 & 1.000 & 1.000 \\
0.9 & 0.35 & 0.050 & 0.993 & 0.961 & 0.100 & 1.000 & 1.000 & 0.200 & 1.000 & 1.000 \\
0.9 & 0.9 & 0.049 & 0.836 & 0.338 & 0.100 & 0.996 & 0.863 & 0.198 & 1.000 & 1.000 \\
0.9 & 0.97 & 0.048 & 0.833 & 0.282 & 0.099 & 0.989 & 0.782 & 0.197 & 1.000 & 1.000 \\
\bottomrule
\end{tabular}
\caption{Detection rates at $\alpha = 0.05$ under mechanism~(a) (\SKpowReps{}
replications per cell).  $\hat r$: realized mean correlation; pooled: pooled
test; comb.: combined criterion (joint form).}
\label{tab:simpower}
\end{table}

\paragraph{Study-level results.}
Table~\ref{tab:simstudy} reports the mean false discovery proportion (FDP;
zero for a study without discoveries) per condition.
Under BH on $p_{\mathrm{final}}$ in its joint form the mean FDP is at most
\SKfdpJointMax\% (approximate 95\% Monte Carlo interval
\SKfdpJointMaxLo--\SKfdpJointMaxHi\%); without true associations none of
the \SKstudies{} studies of either condition has a discovery, and with true
associations the mean true-positive rate is \SKtprJointMin\% or more.
With $B=2{,}000$ the smallest attainable $p_{\mathrm{boot}}$ is
$2/\SKBplusOne$; with \RNnTests{} tests, BH applied directly to the
bootstrap or combined $p$-values therefore yields either zero or at least
three discoveries.
The global-null comparison with the continuous cluster-robust reference is
sensitive to $p$-value resolution.
Its ranking applies to the fixed budget of 2{,}000 resamples.
Of the \SDnullRefLow{} and \SDnullRefHigh{} reference studies with
discoveries in the two conditions without true associations (\SKstudies{}
studies each), \SDnullRefFewLow{} and \SDnullRefFewHigh{} had exactly one or
two, outcomes unavailable under direct BH on the combined $p$-values at this
budget.
The effect of a larger resampling budget on the combined procedure was not
evaluated in the simulation; for the corpus, Section~\ref{sec:results}
reports the analysis with \RBtotalB{} resamples per test.
The same resolution limits detection when few associations are true: in the
two conditions with a probability of 5\%, \SDoneStudies{} stored studies
contain exactly one true association, and the procedure declares it in
\SDoneDeclared{} of them; in the \SDtwoStudies{} studies with exactly two it
declares \SDtwoDeclared{} of \SDtwoTrue{} true associations, and with exactly
three \SDthreeDeclared{} of \SDthreeTrue.
The cluster-robust reference declares all true associations of the studies
with one or two (\SDrefOneDeclared{} and \SDrefTwoDeclared), at a higher mean
FDP in these conditions (Table~\ref{tab:simstudy}).
These subgroups are small and were not fixed in advance; the mean
true-positive rate above does not describe this regime.
The conservative-bound form stays at or below \SKfdpConsMax\%.
The bootstrap alone has a mean FDP of \SKfdpBootMax\% in one condition
(\SKfdpBootMaxLo--\SKfdpBootMaxHi\%; 5\% true associations at $\rho_y=0.35$)
and at most \SKfdpBootSecond\% elsewhere; BH on the cluster-robust reference reaches
\SKfdpCRMax\%, and BH on the pooled $p$-values lies between \SKfdpPoolMin\%
and \SKfdpPoolMax\%.
The screen-then-confirm pipeline of Section~\ref{sec:protocol} does not keep
the mean FDP at 5\%: with the joint criterion it ranges from \SKscJointMin\%
to \SKscJointMax\% and exceeds 5\% in eight of the ten conditions, most
strongly when few or no associations are true.
The pairs it confirms are therefore not a multiplicity-controlled set.
With a separate label series per test the results are nearly the same as
with shared ones.

\begin{table}[t]
\centering
\footnotesize
\begin{tabular}{@{}llr rrrrr r r@{}}
\toprule
 & & & \multicolumn{5}{c}{mean FDP under BH (\%)} & S--C & TPR \\
Labels & $\rho_y$ & True & pooled & joint & cons. & boot. & CR & joint (\%) & joint (\%) \\
\midrule
shared & 0.35 & 0\% & 28.3 & 0.0 & 0.0 & 0.1 & 7.3 & 28.3 & --- \\
shared & 0.35 & 5\% & 15.7 & 3.9 & 2.1 & 5.3 & 6.1 & 15.4 & 93.0 \\
shared & 0.35 & 20\% & 9.2 & 3.7 & 2.1 & 4.5 & 4.6 & 8.4 & 100.0 \\
shared & 0.35 & 40\% & 5.8 & 2.8 & 1.7 & 3.3 & 3.3 & 4.7 & 100.0 \\
shared & 0.6 & 0\% & 73.2 & 0.0 & 0.0 & 0.0 & 6.8 & 68.2 & --- \\
shared & 0.6 & 5\% & 34.5 & 3.6 & 2.3 & 4.8 & 6.0 & 26.8 & 92.2 \\
shared & 0.6 & 20\% & 16.6 & 3.8 & 2.7 & 4.4 & 4.7 & 10.3 & 100.0 \\
shared & 0.6 & 40\% & 9.4 & 2.8 & 2.1 & 3.3 & 3.3 & 5.1 & 100.0 \\
separate & 0.35 & 20\% & 9.1 & 3.5 & 2.0 & 4.4 & 4.5 & 8.2 & 100.0 \\
separate & 0.35 & 40\% & 5.9 & 2.8 & 1.7 & 3.3 & 3.3 & 4.8 & 100.0 \\
\bottomrule
\end{tabular}
\caption{Study-level results (\SKstudies{} studies of \RNnTests{} tests per
condition).  Labels: three label series shared by the tests, or a separate
series per test; True: probability with which each test carries a true association ($r=0.10$).
Mean false discovery proportion under BH on the pooled $p$-values, on
$p_{\mathrm{final}}$ in its joint and its conservative-bound form, on the
bootstrap $p$-value alone, and on a cluster-robust reference (CR); S--C:
screen-then-confirm with the joint form; TPR: mean true-positive rate under
BH on the joint form.}
\label{tab:simstudy}
\end{table}

\paragraph{Outcome of the evaluation rule.}
The joint form meets both conditions: no null cell above \SKtol{} and an
observed mean FDP of at most 5\% in every study condition.
It therefore remains the main procedure of this paper.
The conservative-bound form also meets both, with rejection rates far below
the nominal level in many cells.
The bootstrap alone meets condition~(N) and misses the point threshold of
condition~(S) in one study condition; the cluster-robust reference meets (N)
and exceeds the threshold of (S) in the conditions with no or few true
associations.
These statements hold for the simulated mechanisms.

\paragraph{Scope.}
The simulation uses Gaussian AR(1) and binary Markov building blocks,
complete series for every test, one base rate and one persistence level for
all labels of a condition, true associations of equal size, and conversations
as independent units.
No conversation-specific random effects or user-level clustering are
simulated.

\paragraph{Release.}
The simulation code with its fixed seeds, the per-cell counts of all
procedures, the $p$-values of all simulated studies, the evaluation output,
and a script that recomputes Tables~\ref{tab:simnull}--\ref{tab:simstudy}
from these files ship in the supplement.

\section{AnnoMI Interval Comparison: Protocol and Complete Results}
\label{app:public-extension}

\paragraph{Source and observation units.}
AnnoMI contributes 4{,}817 client and 4{,}882 therapist utterances.
Utterances are ordered by their original identifiers within each transcript
and role; a lag refers to successive utterances of the same role.
Valid client categories are change, sustain, and neutral; valid therapist
categories are question, reflection, therapist input, and other.
The three contrasts are change talk and sustain talk against the other valid
client labels, and reflection against the other valid therapist labels.
A missing or unknown label stays missing at its original position.
Source-video identifiers retain the connection between transcripts from the
same recording.

\paragraph{Eight features.}
Lowercased text is tokenized with the regular expression
\verb|[a-zA-Z']+|.  Let $w_t$ be the token count and $z_t=\log(1+w_t)$.
The features are $z_t$; the number of distinct tokens divided by $w_t$
(undefined when $w_t=0$); trailing means of $z_t$ over complete windows of 5
and 20 observations; EWMA with $\alpha=0.3$ initialized at $z_1$;
the cumulative mean and the cumulative sum of $z_t$; and $z_t-z_{t-1}$.
Every transformation restarts within its transcript and role.
The features use no learned model; they represent construction classes, not
semantic or clinical measurements.

\paragraph{Autocorrelation.}
For the finite positions $F$ of a sequence with mean $\bar x$, the lag-1
estimate is
$\widehat\rho_1=\sum_{t:\,t,t+1\in F}(x_t-\bar x)(x_{t+1}-\bar x)/\sum_{t\in F}(x_t-\bar x)^2$,
with original positions preserved across gaps; at least ten finite values
and five finite lag pairs are required, and undefined estimates are omitted
from means.
Change-talk autocorrelation is defined for 99 of the 133 client sequences,
sustain talk for 86, and reflection for 101 of the 133 therapist sequences.

\paragraph{Interval target.}
All eight features are paired with each of the three contrasts.
Every position with a finite feature value and a valid binary label is
retained.
The statistic is the pooled Pearson correlation, equivalent to the
point-biserial correlation for these labels; longer recordings contribute
more pairs.
A pair is eligible with at least 30 contributing videos and at least 20
positive and 20 negative labels; all 24 pairs qualify.

\paragraph{Video and utterance resampling.}
For video resampling, each replicate draws as many source videos with
replacement as contribute to the pair and pools their complete contributions,
including all transcripts of a selected video; 72--119 videos contribute,
depending on the feature's available positions.
For the utterance-level reference, each replicate draws $n$ of the valid
feature--label pairs with replacement; the already computed features are held
fixed.
Both procedures use 19{,}999 replicates and the 2.5th and 97.5th percentiles
with linear interpolation (seeds 20261117 and 20261118).
No replicate of either procedure is undefined.
A Fisher interval,
$\tanh\{\operatorname{atanh}(r)\pm z_{0.975}/\sqrt{n-3}\}$, is reported as an
additional reference on the same pairs.
All intervals are pointwise, without a multiple-comparison adjustment or a
coverage guarantee.
In Table~\ref{tab:public-intervals}, ``IID'' denotes the utterance-level
resampling.

\par\ifdim\dimexpr\pagegoal-\pagetotal\relax<25\baselineskip \newpage\fi
\begingroup
\scriptsize
\setlength{\tabcolsep}{3pt}
\begin{longtable}{lrrrccc r}
\caption{All 24 AnnoMI control pairs. $n$ counts valid utterance--label pairs;
$k$ counts contributing source videos. Each interval is pointwise 95\%.
IID and video intervals use the same percentile construction with 19,999 draws.
The last column is video/IID interval width, computed before rounding
the displayed endpoints; no multiplicity-adjusted
discovery or false-positive count is inferred.}\label{tab:public-intervals}\\
\toprule
Measure & $n$ & $k$ & $r$ & Fisher & IID percentile & Video percentile & Width ratio\\
\midrule
\endfirsthead
\toprule
Measure & $n$ & $k$ & $r$ & Fisher & IID percentile & Video percentile & Width ratio\\
\midrule
\endhead
\bottomrule
\endfoot
\multicolumn{8}{l}{\textbf{Client: change talk}}\\*
Log length & 4817 & 119 & 0.238 & [0.211, 0.265] & [0.212, 0.264] & [0.190, 0.287] & 1.862 \\
TTR & 4817 & 119 & -0.170 & [-0.197, -0.142] & [-0.199, -0.141] & [-0.214, -0.127] & 1.511 \\
Rolling 5 & 4287 & 117 & 0.139 & [0.110, 0.169] & [0.110, 0.168] & [0.080, 0.203] & 2.102 \\
Rolling 20 & 2713 & 72 & 0.053 & [0.016, 0.091] & [0.016, 0.092] & [-0.048, 0.156] & 2.680 \\
EWMA & 4817 & 119 & 0.193 & [0.166, 0.220] & [0.167, 0.219] & [0.137, 0.250] & 2.146 \\
Cumulative mean & 4817 & 119 & 0.142 & [0.114, 0.169] & [0.115, 0.169] & [0.075, 0.207] & 2.435 \\
Cumulative sum & 4817 & 119 & 0.114 & [0.086, 0.142] & [0.082, 0.145] & [-0.033, 0.221] & 4.005 \\
First difference & 4684 & 119 & 0.136 & [0.108, 0.164] & [0.109, 0.165] & [0.105, 0.171] & 1.170 \\
\midrule
\multicolumn{8}{l}{\textbf{Client: sustain talk}}\\*
Log length & 4817 & 119 & 0.217 & [0.190, 0.244] & [0.191, 0.243] & [0.179, 0.257] & 1.506 \\
TTR & 4817 & 119 & -0.151 & [-0.179, -0.124] & [-0.182, -0.122] & [-0.197, -0.110] & 1.444 \\
Rolling 5 & 4287 & 117 & 0.072 & [0.042, 0.102] & [0.044, 0.101] & [0.033, 0.114] & 1.415 \\
Rolling 20 & 2713 & 72 & -0.036 & [-0.074, 0.001] & [-0.071, -0.001] & [-0.094, 0.017] & 1.587 \\
EWMA & 4817 & 119 & 0.144 & [0.116, 0.172] & [0.117, 0.172] & [0.101, 0.189] & 1.596 \\
Cumulative mean & 4817 & 119 & 0.080 & [0.052, 0.108] & [0.050, 0.110] & [0.031, 0.132] & 1.675 \\
Cumulative sum & 4817 & 119 & -0.062 & [-0.090, -0.034] & [-0.087, -0.036] & [-0.125, -0.033] & 1.776 \\
First difference & 4684 & 119 & 0.127 & [0.099, 0.155] & [0.099, 0.154] & [0.100, 0.155] & 0.989 \\
\midrule
\multicolumn{8}{l}{\textbf{Therapist: reflection}}\\*
Log length & 4882 & 119 & 0.220 & [0.193, 0.247] & [0.197, 0.243] & [0.177, 0.261] & 1.831 \\
TTR & 4881 & 119 & -0.067 & [-0.095, -0.039] & [-0.094, -0.040] & [-0.103, -0.032] & 1.323 \\
Rolling 5 & 4352 & 117 & 0.044 & [0.015, 0.074] & [0.016, 0.073] & [-0.003, 0.091] & 1.671 \\
Rolling 20 & 2753 & 73 & 0.012 & [-0.025, 0.050] & [-0.024, 0.049] & [-0.056, 0.082] & 1.875 \\
EWMA & 4882 & 119 & 0.080 & [0.052, 0.108] & [0.055, 0.105] & [0.032, 0.128] & 1.895 \\
Cumulative mean & 4882 & 119 & 0.003 & [-0.025, 0.031] & [-0.024, 0.030] & [-0.057, 0.065] & 2.239 \\
Cumulative sum & 4882 & 119 & 0.058 & [0.030, 0.086] & [0.028, 0.087] & [-0.024, 0.112] & 2.316 \\
First difference & 4749 & 119 & 0.211 & [0.183, 0.238] & [0.185, 0.236] & [0.176, 0.244] & 1.347 \\
\end{longtable}
\endgroup

\paragraph{Screen-based summary.}
The pooled screen of Section~\ref{sec:annomi-public} uses two-sided $t$-tests
of the pooled correlation with BH over all 24 pairs at $q=0.05$; it retains
21 pairs, of which 17 have a video-level interval that excludes zero.

\paragraph{Reproducibility.}
The supplement provides the numerical input sequences and labels, the
preparation and analysis code, the protocols with their dates, and complete
outputs; raw dialogues are omitted and can be reconstructed from the public
original.

\section{Metric Definitions}
\label{app:metrics}

This appendix defines every metric in the test battery, following the
released analysis code; the battery comprises \textbf{66 metrics} (one of
them, \texttt{Psi}, is an affine transform of \texttt{Entropy}), tested
against up to three labels each, which yields the \RNnTests{} metric--label
tests of the inventory.
A small number of engine parameters are not distributed as numeric
configuration values for responsible-disclosure reasons.  This appendix names
each withheld parameter and marks it ``(RD)''.\enlargethispage{-\baselineskip}
Literal constants that appear directly in the analysis code (window
lengths of standalone functions, the EWMA $\alpha$, ensemble weights and
thresholds) are not withheld and are stated.
Withheld parameters enter 29 of the 66 metrics; the supplement lists the 37
computed without them.
The released inventory allows the tables of this paper to be recomputed from
the stored test results, whereas recomputing those 29 metrics requires the
external configuration.

\paragraph{Autocorrelation estimator.}
Lag-1 autocorrelation is computed within each conversation and averaged
across conversations.
For each metric--label test, only the jointly available observations are
retained, in their original order; lag one therefore refers to neighboring
retained observations, which for the metrics defined only at event onsets
are successive events and not successive turns.
For a nonconstant retained sequence of length $n\geq4$, the estimator is
\[
\widehat\rho_1=\frac{n}{n-1}
\frac{\sum_{t=1}^{n-1}(x_t-\bar x)(x_{t+1}-\bar x)}
{\sum_{t=1}^{n}(x_t-\bar x)^2}.
\]
Shorter or constant metric sequences contribute zero to the mean, a
computational convention, and conversations without a retained observation
do not enter it; for labels, the joint correction averages over the
conversations in which the estimate is defined (Section~\ref{sec:chelton}).

\paragraph{Label-fitted metrics.}
Two metrics, \texttt{hooke\_residual} and \texttt{nyp\_mech}, derive from a
stress--strain fit whose fitting rows are the turns at which both the
\textsc{malicious} and the \textsc{typx} label are available and equal to
zero.
Their associations with these labels are therefore descriptive and not
independent evidence; Section~\ref{sec:results} and Table~\ref{tab:variants}
report the main result without their four tests.
In the rare case of an open yield gate without an evaluable window, the
original fallback leaves a numerical zero also in the main construction;
such zeros are not measurements of absent deformation or absent yield.
\texttt{E\_local} and \texttt{E\_ratio} fit FID against $1-\mathrm{WD}$ on
local or expanding windows without any label-based selection.

\subsection{Preliminaries}

\paragraph{Input quantities.}
The measurement unit is the turn pair: a consecutive (user prompt,
assistant response) pair; turn positions $t$ are counted from zero, as in
the code.
User prompts longer than 200 words are flagged as pasted material and
excluded from the correlation analyses.
Three embedding variants are computed with the sentence transformer
\texttt{all-MiniLM-L6-v2} (384 dimensions): \emph{combined} =
\texttt{encode(prompt + response)} per turn pair, \emph{prompt} =
\texttt{encode(prompt)}, and \emph{response} = \texttt{encode(response)}.
Token weights are word counts of prompt plus response.
Timestamp-derived metrics use the available export timestamps; turns without a usable timestamp are masked.

\paragraph{Embedding normalization.}
The encoder returns unit-norm vectors (verified in the analysis
environment: $\|e\| = 1.0$), so for two embeddings
$\|a - b\|^2 = 2\,(1 - \cos(a,b))$: Euclidean sentence velocity is a
monotone transform of cosine distance.

\paragraph{Two accumulator configurations.}
Two parameterizations of the same budget accumulator run in parallel, a
\textsc{production} and a \textsc{detection} configuration.
\texttt{FID}, \texttt{FID\_weighted}, \texttt{V\_weighted},
\texttt{L\_weighted} derive from \textsc{production};
\texttt{det\_B\_t}, \texttt{det\_c\_t}, and the inputs of
\texttt{B\_windowed} and \texttt{tau\_recovery} derive from
\textsc{detection}.
The withheld parameters (RD) are: \texttt{epsilon} (recurrence threshold),
\texttt{min\_gap}, \texttt{theta\_V}, \texttt{theta\_L},
\texttt{recovery\_rate}, the impulse weights
\texttt{w\_V}/\texttt{w\_L}/\texttt{w\_P}, \texttt{window\_size},
\texttt{theta\_FID}, and \texttt{W\_fid}.
The released configuration interface also requires \texttt{tau\_decay} and
\texttt{Theta\_1}/\texttt{Theta\_2}/\texttt{Theta\_3}; these fields are not used by the
metric-computation path reported here.

\paragraph{Accumulator inputs.}
For the current embedding window, let $a_k$ be the token count at position
$k$, $\bar e_a=\sum_k a_k e_k/\sum_k a_k$, and $d$ denote cosine distance.
The weighted variety is
$V_w=\sum_k a_k d(e_k,\bar e_a)^2/\sum_k a_k$.
Weighted loop density is
\[
L_w=\frac{\sum_k a_k\,\mathbf{1}\{\exists j<k:
 k-j>\texttt{min\_gap},\ d(e_k,e_j)\leq\texttt{epsilon}\}}
 {\sum_k a_k}.
\]
Each window position is counted at most once in this recurrence fraction.
Equal weights give the unweighted $V$ and $L$; the released implementation
returns zero for windows of at most one embedding and stabilizes the vector
normalizations of the variety by adding $10^{-10}$ to their norms.
The impulse uses the weighted inputs and
$P_t=L_{w,t}-L_{w,t-1}$, with the preceding loop density initialized to zero.
The window length and recurrence parameters remain symbolic (RD).
When $V_w<\texttt{theta\_V}$ and $L_w>\texttt{theta\_L}$, the impulse is
\[
c_t=\texttt{w\_V}(1-V_w)+\texttt{w\_L}L_w+
\texttt{w\_P}\max(0,P_t),
\]
subject also to $\mathrm{FID}_{\mathrm{rolling}}>\texttt{theta\_FID}$
when $\texttt{theta\_FID}<1$.  If the applicable gate conditions fail,
$c_t=0$.
Here $\mathrm{FID}_{\mathrm{rolling}}$ is the mean of the last
\texttt{W\_fid} unweighted FID values once that window is full, and the
current FID beforehand.

\paragraph{Design family versus table category.}
Each definition below names the computational design family; the category
used in Tables~\ref{tab:autocorrelation} and~\ref{tab:main} is the coarser
archival grouping, and the two do not coincide for every metric: several
windowed quantities, for instance, are archived under thematic categories
such as frame distance or thermodynamic. Rolling aggregates have mixed window lengths
(\texttt{PRO\_rolling}: 5), and the impulse category includes the binary
\texttt{phase\_collapse} indicator. \texttt{G\_shear} is archived under
lexical/structural despite being an interaction. Neither of the two enters the
\textsc{malicious} BH screen, so their assignment does not affect the counts of
Table~\ref{tab:main}.
The family names follow the dated definitions file: \texttt{LVSK},
\texttt{epsilon\_wd\_rel}, and \texttt{Psi} are listed as per-turn scalars although
they inherit window or whole-conversation dependence from their inputs.
Several metrics use quantities of the whole conversation (the projection and
class limits of \texttt{Entropy}, \texttt{VE}, and \texttt{Psi}; the baselines,
fits, and normalizations of \texttt{epsilon\_wd\_rel}, \texttt{hooke\_residual},
\texttt{nyp\_mech}, \texttt{tau\_recovery}, and \texttt{phase\_collapse}) or later
turns (the recovery lengths); they are retrospective descriptions of a
recorded conversation.

\paragraph{Clarifications.}
(a)~The \texttt{MAS} semantic axis is constructed from fixed external
literal seed sentences using the fixed encoder, then cached per process.
No conversation embeddings or labels are used to fit the axis; conversation
response embeddings enter only its projection step.
(b)~\texttt{sem\_velocity} and its derivatives use the Euclidean
sentence-velocity function, not the cosine-based variant inside the budget
accumulator.
(c)~Inside \texttt{drift\_score}, the weave-density contribution is the
proxy $L\,(1-V)$. Its internal \texttt{NC2} input is fixed at zero,
which makes that transformed sub-score one and its weighted contribution
a constant $0.10$; it is not the effective-rank column \texttt{NC2}.
(d)~\texttt{phase\_collapse} depends only on the normalized weave-density
slope: the variety series passed to the released function enters only the
other phase labels, which are not part of the battery, and the
frame-distance series that the original function also received was unused.
(e)~\texttt{sem\_velocity\_t} and \texttt{turn\_pressure} follow the timestamp masks; a finite value alone does not verify the provenance of its time axis.

\subsection{Definitions}

The definitions describe the released base code unless a rule of the main
construction is named; the availability rules of the main construction
apply in addition (end of this appendix).

\paragraph{Per-turn scalars.}
\begin{itemize}[nosep]
\item \textbf{FID} (frame distance): $\max(0,\, 1 - \cos(e_t,
  C_{\mathrm{frame}}))$ between the combined turn embedding and the frame
  centroid $C_{\mathrm{frame}}$ = mean of the first \texttt{window\_size}
  (RD) combined embeddings (\textsc{production}).  Drift from the opening
  topic.
\item \textbf{DIV}: $1 - \cos(e^{\mathrm{prompt}}_t,
  e^{\mathrm{response}}_t)$, clipped at 0.  Prompt--response semantic
  mismatch.
\item \textbf{QR}: question density of the user prompt,
  $\min(1, (\#\text{question marks} + \#\text{question words}) /
  \#\text{sentences})$.
\item \textbf{MR}: meta-reference rate of the response (lexicon-based),
  $\min(1, \#\text{meta markers}/\#\text{tokens})$.
\item \textbf{MAS}: cosine projection of the L2-normalized response
  embedding onto a bipolar meta-awareness axis
  $\mathrm{norm}(c_{\mathrm{pos}} - c_{\mathrm{neg}})$ built from fixed
  seed sentences (meta-reflection vs.\ factual assertion).
\item \textbf{LVSK}: type-token ratio of the last up to 100 tokens of the
  current prompt--response pair, divided by the unweighted variety $V$ of
  the \textsc{production} accumulator (RD upstream) when $V\geq10^{-6}$ and
  by $0.001$ otherwise; with fewer than five tokens the base code returns
  0.  High = lexically varied but semantically converged.
\item \textbf{sem\_velocity}: Euclidean step length
  $\|e_t - e_{t-1}\|$ of consecutive combined embeddings.
\item \textbf{sem\_velocity\_w}: token-normalized step length
  $\|e_t - e_{t-1}\| / (\mathrm{tok}_t + \mathrm{tok}_{t-1}) \times 100$
  (movement per 100 tokens).
\item \textbf{dir\_consistency}: at turn $t$,
  $\cos(e_{t-1}-e_{t-2},\,e_t-e_{t-1})$, the cosine of the two most recent
  step vectors of the combined embeddings, evaluated where both step norms
  exceed $10^{-9}$ (the code returns 0 otherwise).  High = straight-line
  drift.
\item \textbf{NCD}: normalized compression distance (gzip) between prompt
  and response of the same turn pair,
  $(C(xy) - \min(C(x), C(y)))/\max(C(x), C(y))$.  Byte-level, orthogonal
  to embeddings.
\item \textbf{epsilon\_wd\_rel}: relative strain
  $(\mathrm{WD}_{\mathrm{base}} - \mathrm{WD}(t)) /
  \mathrm{WD}_{\mathrm{base}}$ with $\mathrm{WD}_{\mathrm{base}}$ = median
  of the first $\max(5, \min(20, \lfloor T/5\rfloor))$ weave-density values;
  where $\mathrm{WD}_{\mathrm{base}}\leq10^{-6}$ the code uses
  $1-\mathrm{WD}(t)$.
\item \textbf{Psi}: order parameter $1 - \mathrm{Entropy}/\log 5$ (see
  \textbf{Entropy}).
\end{itemize}

\paragraph{Rolling aggregates.}
\begin{itemize}[nosep]
\item \textbf{WD} (weave density): share of the unordered pairs among the
  combined embeddings at positions $t-5,\ldots,t$, a six-embedding window
  (code argument \texttt{window=5}), with cosine distance $\leq$
  \texttt{epsilon} (RD).  The base code returns 0 before this window is
  available; the main construction treats these positions as missing.
  Recurrence density.
\item \textbf{ARI\_rolling}: rolling mean ($W{=}20$) of
  $\mathrm{ARI}[t] = \max(0, (\mathrm{DIV} - \mathrm{VS})/2)$, where VS is
  a lexicon-based valence of the response.
\item \textbf{E\_Modul}: stiffness $1/(\operatorname{Var}(\mathrm{VE}
  \text{ over } W{=}20) + 10^{-6})$.
\item \textbf{Entropy}: Shannon entropy (nats) of the response classes over
  a trailing window of up to $W{=}5$ turns.  Classes: prompt and response
  embeddings are projected onto the first principal component fitted to all
  prompt and response embeddings of the conversation and assigned to 5
  classes at the quantiles of these projections; projection and class
  limits therefore use the whole conversation and are not refitted per
  window.
\item \textbf{VE}: plug-in mutual information (nats) between the prompt
  and response classes of the same projection (1-D PCA, 5 classes) over the
  same trailing window.
\item \textbf{F\_CV}: coefficient of variation (standard deviation over
  mean) of the last up to five values of the folding energy $F_t$, where
  $F_t$ is the variance of the cosine distances of the last up to five
  combined embeddings to their centroid.  Each step needs at least three
  values; otherwise, and for a mean of at most $10^{-10}$, the code
  returns 0.
\item \textbf{NC2}: effective rank (participation ratio) of the centered
  cloud of combined embeddings over a trailing window of up to $W{=}20$
  turns, evaluated from five embeddings on: $1/\sum_i p_i^2$ with
  $p_i = \sigma_i / \sum \sigma$ over the singular values above $10^{-10}$
  (undefined with fewer than two).  Low = collapsed cloud.
\item \textbf{PRO\_rolling}: per turn, the number of relational pronouns
  divided by one plus the number of factual pronouns in prompt and response
  (multilingual lexica).  Conversations of at most five turns retain these
  ratios; longer conversations use their trailing mean over up to five
  turns.
\item \textbf{Hysteresis\_rolling}: for the $m\leq10$ most recent
  unweighted $(L,V)$ values of the \textsc{production} accumulator, the
  mean of the $m-1$ terms $\tfrac{1}{2}(L_k+L_{k+1})\,|V_{k+1}-V_k|$ (RD
  upstream); the base code returns 0 for $m\leq1$.
\item \textbf{drift\_score}: fixed-weight ensemble
  $0.10 + 0.35\,s_L + 0.30\,s_{L(1-V)} + 0.15\,s_F + 0.10\,s_V$
  with $s_L=\min(1,L/0.84)$, $s_{L(1-V)}=\min(1,L(1-V)/0.70)$,
  $s_F=\max(0,1-F_{\mathrm{fold}}/2.10)$, and $s_V=\max(0,1-V/0.16)$, where
  $F_{\mathrm{fold}}=-\log(d+10^{-9})$ and $d$ is the share of pairs of
  combined embeddings in a trailing window of up to five turns with cosine
  distance at most \texttt{epsilon} (RD); for a window of a single
  embedding the code sets $F_{\mathrm{fold}}=0$.  The four variable
  contributions have total weight $0.90$, without renormalization; $0.10$
  is a constant offset (clarification~(c)).
\item \textbf{EWMA\_sem\_velocity}: exponentially weighted moving average of
  \texttt{sem\_velocity}, $\alpha = 0.3$ (literal).
\item \textbf{rolling\_predictability}: $R^2$ of a linear fit of the
  EWMA velocity over the preceding $W{=}5$ turns; falling $R^2$ indicates a change in the local linear-fit statistic independent of direction.
\item \textbf{wd\_oscillation}: rolling variance of WD over $W{=}10$.
\item \textbf{B\_windowed}: windowed budget
  $\max(0,\sum_{s\in I_t}c_s-|I_t|\cdot\texttt{recovery\_rate})$ with
  $I_t=\{\max(0,t-19),\ldots,t\}$, on the \textsc{detection} impulses and
  recovery rate (RD).
\end{itemize}

\paragraph{Cumulative and normalized-cumulative metrics.}
\begin{itemize}[nosep]
\item \textbf{det\_B\_t}: accumulated budget
  $B_t=\max(0,B_{t-1}+c_t-\texttt{recovery\_rate})$, initialized at zero,
  using the impulse defined above under the \textsc{detection}
  configuration (parameters RD). Palmgren--Miner-style damage accumulation
  over turns.
\item \textbf{Toughness}: cumulative sum of the VE series.
\item \textbf{Toughness\_rate}: the cumulative sum of VE divided by the
  number of summands: $\mathrm{Toughness}[t]/(t+1)$ in the base code,
  which includes its initialized opening value; the main construction
  restarts the sum after the excluded opening turn and divides by the
  number of retained summands.
\end{itemize}

\paragraph{First differences / impulses.}
\begin{itemize}[nosep]
\item \textbf{det\_c\_t}: the \textsc{detection} instability impulse
  $c(t)$ itself (parameters RD).
\item \textbf{delta\_WD, delta\_FID, delta\_PRO, delta\_NC2,
  delta\_NCD}: first differences $x_t - x_{t-1}$.  The opening difference
  of a conversation is undefined and is treated as missing in the main
  construction.
\end{itemize}

\paragraph{Interaction terms.}
\begin{itemize}[nosep]
\item \textbf{coherence\_trap}: product $\mathrm{AR}(t) \cdot
  \mathrm{FID}(t)$, where AR is an agreement-marker rate of the response
  (lexicon-based, $\min(1, \#\text{agreement markers}/\#\text{words})$).
\item \textbf{coherence\_trap\_weighted}: $\mathrm{AR}(t) \cdot
  \mathrm{FID\_weighted}(t)$.
\item \textbf{WD\_x\_FCV}: $\mathrm{WD}\,(1 - \mathrm{F\_CV})$.
  \textbf{WD\_div\_FCV}: $\mathrm{WD}/\mathrm{F\_CV}$ (NaN below
  $10^{-6}$).
\item \textbf{G\_shear}: $\mathrm{DIV}/|\Delta \mathrm{WD}|$ (NaN below
  $10^{-6}$); archived under lexical/structural, computationally an
  interaction term.
\item \textbf{NCD\_x\_DIV}: $\mathrm{NCD} \cdot \mathrm{DIV}$.
\item \textbf{stiffness}: $\mathrm{dir\_consistency} /
  (\mathrm{sem\_velocity} + 10^{-6})$.
\item \textbf{E\_t}: elasticity operator: the mean of the five preceding
  first differences of direction consistency, divided by the absolute mean
  of the corresponding first differences of velocity plus $10^{-6}$.
\item \textbf{E\_ratio}: $E_{\mathrm{local}}/E_{\mathrm{global}}$,
  where $E_{\mathrm{global}}$ fits FID against $1-\mathrm{WD}$ on the
  preceding expanding window (starting after the initialization turns).
  Neither fit selects rows using target labels.
\end{itemize}

\paragraph{Rolling regression and residual metrics.}
\begin{itemize}[nosep]
\item \textbf{E\_local}: local slope (degree-1 fit) of FID against strain
  $\varepsilon = 1 - \mathrm{WD}$ over sliding $W{=}5$.
\item \textbf{hooke\_residual}: $|\sigma(t) - (E_{\mathrm{eff}}\,
  \varepsilon(t) + b)|$ with $\sigma$ = FID, $\varepsilon$ = relative
  strain, and $(E_{\mathrm{eff}}, b)$ fitted on the non-pasted turns at which both labels are available and 0 and $|\varepsilon|>10^{-6}$ (at least five such turns with a strain standard deviation above $10^{-8}$; otherwise the base code returns 0). Deviation from the linear stress--strain relation; because the fit uses the labels, its association with those labels is not an independent prediction.
\end{itemize}

\paragraph{Binary state indicators.}
\begin{itemize}[nosep]
\item \textbf{nyp\_mech}: 1 from the first of three consecutive turns at
  which the rolling stress--strain $R^2$ lies below $0.3\times$ the global
  $R^2$ of the fit of \texttt{hooke\_residual} (mechanical yield-point
  indicator).  A rolling fit uses the ten preceding positions, restricted
  to non-pasted turns with both labels available (of either value) and
  $|\varepsilon|>10^{-6}$, and needs at least five such turns with a strain
  standard deviation above $10^{-8}$.  The
  indicator is evaluated only for conversations of more than 15 turns with
  a global $R^2>0.05$; otherwise the base code returns 0.
\item \textbf{phase\_collapse}: 1 when the normalized WD slope falls
  below $-0.5$ (sharp weave-density drop): the least-squares slope of the
  five preceding WD values, divided by the larger of $10^{-6}$ and the
  standard deviation of these slopes over the conversation.  The base code
  copies the value of the first evaluated turn to the opening turns.
  Archived under impulse, computationally a binary indicator
  (clarification~(d)).
\item \textbf{neg\_elasticity}: 1 at a turn whose first difference of
  direction consistency lies below $-\sigma_{dc}$ while that of velocity
  lies above $\sigma_{sv}$, where each $\sigma$ is the standard deviation
  of the ten preceding first differences of the same series (plus
  $10^{-9}$).
\end{itemize}

\paragraph{Event/recovery metrics.}
\begin{itemize}[nosep]
\item \textbf{recovery\_len}: at each perturbation onset, the number of
  turns until WD re-enters its pre-perturbation band (mean $\pm 1\sigma$,
  lookback 5; cap 20); set at the onset turn, else NaN.  A candidate spike
  at turn $t$ requires both $s_t>0$ and
  $|\Delta\mathrm{WD}_t|>1.5\,s_t$, where $s_t$ is the standard deviation
  of the ten preceding absolute first differences; consecutive candidates
  form one onset at their largest value, and an onset closer than three
  turns to the previous one is dropped.
\item \textbf{recovery\_trend}: slope of the last three recovery lengths
  against their ordinal index (latest minus third latest, divided by two),
  set at the onset turn from the third event on.
\item \textbf{recovery\_len\_sv}: as \texttt{recovery\_len} on the
  velocity series.
\item \textbf{tau\_recovery}: turns until \texttt{det\_B\_t} returns
  below baseline $+1\sigma$ after a spike above baseline $+2\sigma$, set
  at the spike turn (baseline = median of the first
  $\max(5,\lfloor 0.2\,T\rfloor)$ values; $\sigma$ = standard deviation of
  the whole series plus $10^{-9}$; without a return after a spike at turn
  $t$, the recorded value is $T-t$; undefined for $T<10$; RD upstream).
\end{itemize}

\paragraph{Timestamp-based metrics.}
\begin{itemize}[nosep]
\item \textbf{sem\_velocity\_t}: $\mathrm{sem\_velocity} / \log(1 +
  \max(\Delta t, 0) + 10^{-10})$ with $\Delta t$ the timestamp difference;
  undefined where no usable timestamp is available.
\item \textbf{turn\_pressure}: number of turns within the preceding 300
  seconds, the current one included (burst density); undefined where no
  usable timestamp is available.
\end{itemize}

\paragraph{Token-weighted core variants.}
\begin{itemize}[nosep]
\item \textbf{V\_weighted}: token-weighted semantic variety (weighted
  mean of squared cosine distances to the token-weighted window centroid;
  \texttt{window\_size} RD).
\item \textbf{L\_weighted}: token-weighted loop density (weighted share
  of recurrent points under \texttt{epsilon}/\texttt{min\_gap}, RD).
\item \textbf{WD\_weighted}: the six-embedding recurrence fraction of
  \texttt{WD} with pair weight $(a_i+a_j)/2$ and denominator
  $\max(\text{sum of pair weights},10^{-10})$ (\texttt{epsilon} RD).
\item \textbf{FID\_weighted}: FID against the token-weighted frame
  centroid (\texttt{window\_size} RD).
\end{itemize}

\paragraph{Role-separated user/AI metrics.}
\begin{itemize}[nosep]
\item \textbf{sigma\_user}: $\mathrm{cosdist}(e^{\mathrm{user}}_t,
  C^{\mathrm{AI}}_{\mathrm{recent}}) \cdot d_{\mathrm{persist}}$, with
  $C^{\mathrm{AI}}_{\mathrm{recent}}$ the centroid of the $W{=}5$ AI
  response embeddings preceding turn $t$ and $d_{\mathrm{persist}} =
  \max(0, 1 + \cos(u_{t-1}-u_{t-2},\,u_t-u_{t-1}))$ for the user
  embeddings $u$; $d_{\mathrm{persist}}=1$ when either step norm is at most
  $10^{-9}$.
\item \textbf{sigma\_user\_base}: as above against the centroid of the
  \emph{first} $W{=}5$ AI embeddings.
\item \textbf{epsilon\_ai}: $\mathrm{cosdist}(e^{\mathrm{AI}}_t,
  C^{\mathrm{AI}}_{\mathrm{base}})$, AI deformation relative to its
  opening state.
\item \textbf{E\_ai}: $\mathrm{sigma\_user}(t) /
  \mathrm{epsilon\_ai}(t)$ where $\mathrm{epsilon\_ai}(t)>10^{-8}$,
  undefined otherwise; stiffness of the user/AI coupling.
\item \textbf{sv\_user, sv\_ai}: cosine step distance of consecutive
  user (resp.\ AI) embeddings.
\item \textbf{asymm}: $(\mathrm{sv\_ai} - \mathrm{sv\_user}) /
  (\mathrm{sv\_ai} + \mathrm{sv\_user} + 10^{-8})$.
\end{itemize}

\subsection{Count and release}

The supplement provides machine-readable definitions (families, categories,
inputs, windows, and disclosure flags), the per-pair inventory, and
validation scripts for the \RNnTests{} metric--label tests.
The released metric-computation code implements the definitions above with
the original initialization rules; without the withheld configuration file,
which is documented by a template, it computes the metrics that need no
withheld parameter.
The main construction (Section~\ref{sec:setup}) applies missing-initialization
masks and rebuilds the affected recursive, cumulative, and fitted quantities
under the rules described in the supplement, which does not list
metric-specific initialization lengths; these reconstruction steps are not
implemented in the released base metric-computation code.

\end{document}